\documentclass[preprint,a4paper]{elsarticle}

\usepackage{graphicx}
\usepackage[fleqn]{amsmath}
\usepackage{amsfonts}
\usepackage{amssymb}
\usepackage{float}
\usepackage[dvips]{epsfig}
\usepackage{epstopdf}
\usepackage{rotating}
\usepackage{epigraph}
\usepackage{setspace}

\usepackage{array}
\newcolumntype{P}[1]{>{\centering\arraybackslash}p{#1}}
\newcolumntype{M}[1]{>{\centering\arraybackslash}m{#1}}
\usepackage{hhline}

\usepackage{microtype}
\usepackage{url}
\usepackage{subfiles}
\usepackage{caption}
\usepackage{subcaption}
\usepackage{parskip}
\usepackage[section]{placeins}
\usepackage{xcolor}
\usepackage{framed}
\colorlet{shadecolor}{blue!20}
\usepackage{tabulary}
\usepackage[hidelinks]{hyperref}
\usepackage[capitalize,nameinlink]{cleveref}
\usepackage{flexisym}

\usepackage{amsmath,amsthm,amssymb}

\usepackage{calrsfs}
\DeclareMathAlphabet{\pazocal}{OMS}{zplm}{m}{n}

\usepackage{multirow, bigdelim}

\usepackage{tikz}
\usetikzlibrary{decorations.pathmorphing}
\usetikzlibrary{decorations.markings}
\usetikzlibrary{arrows.meta,bending}
\usetikzlibrary{arrows.meta}
\usepackage{bm}

\usepackage{adjustbox}

\newcommand\norm[1]{\left\lVert#1\right\rVert}

\DeclareMathOperator{\Cor}{Cor}

\usepackage{makecell}

\journal{Mechanical Systems and Signal Processing}

\begin{document}
	
	\begin{frontmatter}
		
				\title{On the application of generative adversarial networks for nonlinear modal analysis}

    \author[1]{G.\ Tsialiamanis\footnote{Corresponding Author: George Tsialiamanis (g.tsialiamanis@sheffield.ac.uk)}}
    \author[1]{M.D.\ Champneys}
    \author[1]{N.\ Dervilis}
    \author[1]{D.J.\ Wagg}
    \author[1]{K.\ Worden}
    \address[1]{Dynamics Research Group, Department of Mechanical Engineering, University of Sheffield \\ Mappin Street, Sheffield S1 3JD}

		\begin{abstract}
        Linear modal analysis is a useful and effective tool for the design and analysis of structures. However, a comprehensive basis for \textit{nonlinear} modal analysis remains to be developed. In the current work, a machine learning scheme is proposed with a view to performing nonlinear modal analysis. The scheme is focussed on defining a one-to-one mapping from a latent `modal' space to the natural coordinate space, whilst also imposing orthogonality of the mode shapes. The mapping is achieved via the use of the recently-developed \textit{cycle-consistent generative adversarial network} (cycle-GAN) and an assembly of neural networks targeted on maintaining the desired orthogonality. The method is tested on simulated data from structures with cubic nonlinearities and different numbers of degrees of freedom, and also on data from an experimental three-degree-of-freedom set-up with a column-bumper nonlinearity. The results reveal the method's efficiency in separating the `modes'. The method also provides a nonlinear superposition function, which in most cases has very good accuracy.
		\end{abstract}
		
		
	\end{frontmatter}
	
	
\section{Introduction}
\label{sec:introduction}

Many approaches have been followed throughout the years in order to perform dynamic analysis of structures, the most dominant being \textit{modal analysis} \cite{ewins2009modal}. The reason that modal analysis has proved so powerful is because it provides a meaningful \textit{decomposition} of the oscillations of a structure. The components of this decomposition, the \textit{modes}, are independent, and each one has its own dynamic characteristics, i.e. mode shape, natural frequency, modal damping, modal mass, etc; providing the users with independent movements to study, and ways to isolate potential problems while designing a structure. For linear structures, linear modal analysis is by far the most common method of analysis \cite{chopra2007dynamics}. Both in modelling and at the experimental level, modal decomposition of linear structural behaviour has been achieved to a high level of accuracy. The scheme also provides a practical solution for multi-degree-of-freedom (MDOF) systems by decoupling them into single-degree-of-freedom (SDOF) systems, using an eigen-decomposition of the system matrices where the modes are related to eigenvectors and natural frequencies related to the eigenvalues.

Apart from assisting in understanding the resonant behaviour of structures, modal analysis also provides a convenient model reduction technique. Higher-frequency modes absorb less energy from the excitation and so tend to affect the behaviour of the structure less; thus, they can quite often be omitted. Under a such a framework, techniques, such as principal component analysis (PCA) \cite{wold1987principal} and proper orthogonal decomposition (POD) \cite{chatterjee2000introduction}, are similarly followed with a view to identifying vibration modes from available data \cite{poncelet2007output} and to solve by projecting the algebraic systems onto a lower-dimensional vector basis. The aforementioned scheme is followed when one has no information about the excitation and is often referred to as \textit{operational modal analysis} (OMA) or \textit{data-driven} modal analysis.

Despite the fact that existing modal analysis methods are focussed on linear systems, they are often used on structures with suspected nonlinearities. Modal analysis of structures, which may have structural elements with nonlinear behaviour, is often plausible in cases of real structures, for some range of external loads that does not suffice for the nonlineartities to affect the behaviour of the structure. Consider the \textit{Duffing oscillator} \cite{kovacic2011duffing}, which has a cubic term in its differential equation. For small values of excitation force, the system may not exhibit notable nonlinear behaviour, making linear analysis appropriate and sufficient. For MDOF systems with similar nonlinearities, decomposition of the movement into modes may be achieved in similar cases of low-force excitation, if there is a stable underlying linear system; a necessary but not always sufficient criterion. Nevertheless, this is not always the case. Real-life structures exhibit nonlinear behaviour quite often and linear modal analysis methods fail to define nonlinear modes of vibration. Methods have been developed to deal with such issues \cite{WordenNL}, but only achieve preservation of a \textit{subset} of the properties of linear modal analysis \cite{vakakis1997non, mikhlin2010nonlinears}. 

An approach to data-driven nonlinear modal analysis using machine learning was proposed in \cite{worden2017machine} and \cite{dervilis2019nonlinear}. The idea of structural independence, together with a Shaw-Pierre concept \cite{shaw1993normal} ansatz were used to motivate a new definition of nonlinear normal modes (NNMs). In \cite{worden2017machine}, a genetic algorithm was used to define a decomposition of the displacements of various nonlinear systems into a modal space. With the objective function of the algorithm, a type of orthogonality of the modes, via their statistical independence, was enforced. The approach agreed with the POD in the limit of linear behaviour. The results were encouraging and revealed that such an approach can decouple NNMs from nonlinear systems more effectively than conventional linear decomposition methods. Although the approach presented in \cite{worden2017machine} yielded a good modal decomposition for a two-degree-of-freedom simulated system with cubic nonlinearity, the results were not as good for a three-degree-of-freedom system with similar nonlinearity and for an experimental case. Moreover, the best model selection was achieved `by eye', since the objective function of the optimisation problem did not yield the best results in terms of mode separation. Also, the nonlinear superposition function studied in \cite{worden2017machine} did not yield satisfactory results.

In the current work, following the same machine learning framework and defining an NNM using the same assumptions, an alternative approach is followed. Instead of trying to decompose the natural coordinate space into modal coordinates, a mapping between the two is sought, using a predefined modal space. In order to achieve the desired result, a recently-developed algorithm is used, the \textit{cycle-consistent generative adversarial network} (cycle-GAN) \cite{zhu2017unpaired}. The algorithm is used to define a forward mapping from the natural coordinate space to a modal space, as well as the inverse mapping, to achieve (nonlinear) superposition of the modes. A great advantage of the algorithm, as it will be described, is its invertibility property, so one has, as a consequence, a smooth mapping from one space to the other. Furthermore, a second neural network assembly is used that forces the transition from modal to natural coordinates to enforce the orthogonality of the mode shapes and so to satisfy an NNM criterion. Compared to previous approaches, the new method is a way of training both forward and backward mappings at the same time and identifying the mapping that most efficiently separates the structural movement into independent modes.

The layout of the paper is as follows. Section \ref{sec:gans} provides a brief introduction to generative adversarial networks, to problems that arise when using them and to the cycle-GAN algorithm and how it resolves some of these problems. In Section \ref{sec:orthogonality}, the layout of the neural network, which is used in parallel with the cycle-GAN in order to maintain the desired orthogonality, is described. In Section \ref{sec:ModalCycle} the proposed nonlinear modal analysis algorithm is described. In Section \ref{sec:case_studies}, applications to simulated dynamical systems are presented, together with an application to an experimental dataset. Section \ref{sec:superposition} considers the inverse mapping and superposition of the modes. In Section \ref{sec:cor_study} the correlation of the modes is studied for two correlation metrics, a linear and a nonlinear metric. Finally, in Section \ref{sec:conclusions}, conclusions are drawn about the method.

\section{Generative adversarial networks (GAN)}
\label{sec:gans}

\subsection{Vanilla GAN}
\label{sec:vanilla}

A recent approach to machine learning is that of using \textit{generative models}. Such models are able to learn the distribution of given data and generate artificial data according to it. A trivial approach to the problem of generating artificial data would be to define a Gaussian distribution using the mean and the covariance of the data. Following such an approach, the normal distribution could be used to generate artificial data. However, real data distributions are more complicated; for example, to cope with multi-modal distributions, a kernel density estimate \cite{Epanechnikov1969} could be calibrated according to the data.

A more recent approach is to use a \textit{Generative Adversarial Network} (GAN) \cite{goodfellow2014generative}. The algorithm was initially created to generate synthetic images that look real; i.e. the model learns how to embed figures into some latent space and simultaneously how to generate data according to a proper distribution. Apart from the main goal of the algorithm, a novel way of training neural networks was introduced. \textit{Adversarial training} is defined as a competition between two neural networks. In the basic GAN, the first network is the \textit{generator}, which tries to generate samples that look real and the second is the \textit{discriminator}, which tries to identify whether a sample comes from the real dataset or is artificial.

Training is orchestrated as a competition between two networks. The \textit{discriminator} $D$ is a network with an output representing the probability of its input sample $\bm{x}$ being real; i.e. $P_{\bm{x} \sim p_{data}} = D(\bm{x})$. Throughout training, real and fake samples are introduced to the discriminator and using back-propagation it becomes better at distinguishing samples from the real dataset from generated/artificial samples. The discriminator essentially draws a decision boundary around the manifold of the available data. On the other hand, the \textit{generator} $G$ takes as input a noise vector $\bm{z}$ from some pre-defined probability distribution $p_{z}(\bm{z})$ and creates a sample $G(\bm{z})$ in the feature space of the dataset. Thereafter, the sample is passed through the discriminator in order to decide whether it is real or generated. The probability of a generated sample being real is given by $D(G(\bm{z}))$. Forcing the generator to create samples that `fool' the discriminator into classifying them as real (i.e.\ minimisation of $\log (1-D(G(\bm{z})))$), results in creating samples/images that look real. The optimisation problem based on an objective function $V(D, G)$ for the training of both networks is given by,

\begin{equation} \label{eq:obj_fun}
    \min\limits_{G}\max\limits_{D}V(D,G)=\mathbb{E}_{\bm{x} \sim p_{data}(\bm{x})}[\log D(\bm{x})] + \mathbb{E}_{\bm{z} \sim p_{z}(\bm{z})}[\log(1 - D(G(\bm{z})))]
\end{equation}
where $\mathbb{E}_{x \sim p_{data}(\bm{x})}[\ ]$ is the mean value of the functions within the brackets $[\ ]$ with respect to the distribution $p_{data}$.

\begin{figure}[h!]
    \centering
    \begin{adjustbox}{width=0.7\paperwidth,center}
    \begin{tikzpicture}
        \definecolor{blue1}{RGB}{0, 128, 255}
        \node (1) at (0.0, 0.0) [draw, line width=0.5mm, fill=orange] {Noise, $\bm{z}$};
        
        \node (2) at (3.4, 0.0) [draw, line width=0.5mm, fill=blue1, minimum height=1.5cm, minimum width=2cm] {Generator};
        \draw[-{Latex[width=3mm, length=4mm]}, line width=0.5mm] (1) to (2);
        
        \node (3) at (6.8, 1.0) [draw, line width=0.5mm, fill=orange, minimum width=3.0cm] {\shortstack{Generated\\samples $G(\bm{z})$}};
        \node (4) at (6.8, -1.0) [draw, line width=0.5mm, fill=orange, minimum width=3.0cm] {\shortstack{Real\\ samples $\bm{x}$}};
        \draw[-{Latex[width=3mm, length=4mm]}, line width=0.5mm] (4.4, 0.0) to (5.3, 1.0);
        
        \node (5) at (11.2, 0.0) [draw, line width=0.5mm, fill=blue1, minimum height=1.5cm, minimum width=3.0cm] {Discriminator};
        
        \draw[-{Latex[width=3mm, length=4mm]}, line width=0.5mm] (8.3, 1.0) to (9.7, 0.0);
        \draw[-{Latex[width=3mm, length=4mm]}, line width=0.5mm] (8.3, -1.0) to (9.7, 0.0);
        
        \node (6) at (15.8, 0.0) [draw, line width=0.5mm, fill=orange, minimum height=1.5cm, minimum width=3.0cm] {Probability $D(G(\bm{z}))$};
        \draw[-{Latex[width=3mm, length=4mm]}, line width=0.5mm] (5) to (6);
        
    \end{tikzpicture}
    \end{adjustbox}
    \caption{Layout of a basic (vanilla) GAN in full assembly; the generator transforms noise into generated samples and the discriminator attempts to distinguish between real and generated samples.}
    \label{fig:gan_layout}
\end{figure}
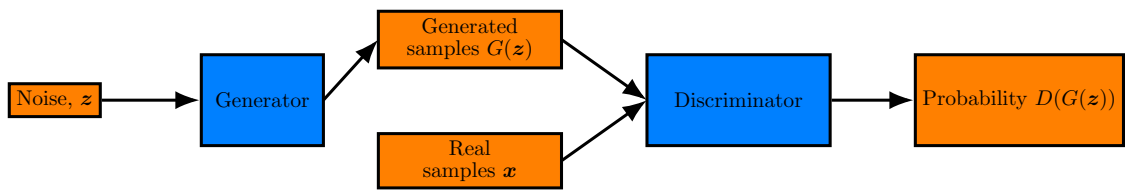

The layout of a basic GAN is shown in Figure \ref{fig:gan_layout}. In practice, training is performed in two steps. During the first step, a batch of samples is randomly sampled from the dataset along with an equally-large batch sampled from the generator. Training of the discriminator \textit{only} is then performed using as target labels $1$ (real) for the dataset samples and $0$ (fake) for the generated samples. During this step, the discriminator is trained to better distinguish, real from fake samples. During the second step, the generator is trained, while the parameters of the discriminator are held constant. For this step, a batch of noise vectors are sampled and the two networks are connected together as shown in Figure \ref{fig:gan_layout}. The second term of equation (\ref{eq:obj_fun}) is used alone for training and the target labels for the output of the discriminator are $1$s, meaning that the generator should transform the noise vectors into samples that the discriminator accepts as real.

GANs in structural dynamics can be used in order to generate artificial data; in cases where acquiring data from structures is expensive, this aspect of GANs is useful and reduces the cost of recording data. GANs may also be used in other ways, to learn mappings from one space to another. A simple approach would be to learn a mapping from a latent space to a high-dimensional space of data. One would hope that via such mappings, each latent variable would encode distinct features of the data, but this is not usually the case; GANs tend to provide an entangled representation of the data via their latent variables.

\begin{figure}[H]
    \centering
    \begin{tikzpicture}[scale=0.65, every node/.style={transform shape}]
        
        \draw[thick] (-2,0) -- (-2,2);
        
        \draw[thick] (-2, 2) -- (-2.2, 1.8);
        \draw[thick] (-2, 1.8) -- (-2.2, 1.6);
        \draw[thick] (-2, 1.6) -- (-2.2, 1.4);
        \draw[thick] (-2, 1.4) -- (-2.2, 1.2);
        \draw[thick] (-2, 1.2) -- (-2.2, 1.0);
        \draw[thick] (-2, 1.0) -- (-2.2, 0.8);
        \draw[thick] (-2, 0.8) -- (-2.2, 0.6);
        \draw[thick] (-2, 0.6) -- (-2.2, 0.4);
        \draw[thick] (-2, 0.4) -- (-2.2, 0.2);
        \draw[thick] (-2, 0.2) -- (-2.2, 0.0);
        
        \draw[thick, decoration={aspect=0.65, segment length=3mm,
             amplitude=0.2cm, coil}, decorate] (-2,1) --(0,1);
        
        \draw[thick] (0,0) rectangle (2,2) node[pos=.5] {$m_1$};
        
        \draw[thick, decoration={aspect=0.65, segment length=3mm,
             amplitude=0.2cm, coil}, decorate] (2,1) --(4,1);
             
        \draw[thick] (4,0) rectangle (6,2) node[pos=.5] {$m_2$};
        
        \draw[thick, decoration={aspect=0.65, segment length=3mm,
             amplitude=0.2cm, coil}, decorate] (6,1) --(8,1);
        
        \draw[thick] (8,0) rectangle (10,2) node[pos=.5] {$m_3$};
        
        \draw[->,thick] (1, 2) -- (1, 3) -- (2, 3);
        
        \node[] at (1.5, 3.5) {$F$};
        
        \node[] at (-1, 1.7) {$k_1$};
        \node[] at (3, 1.7) {$k_2$};
        \node[] at (7, 1.7) {$k_3$};
        
        \end{tikzpicture} 
    \caption{Three-degree-of-freedom mass spring system.}
    \label{fig:mass_spring_example}
\end{figure}
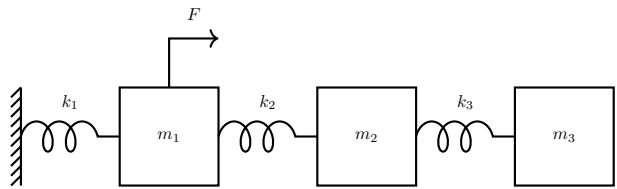

\begin{figure}[H]
\centering
    \begin{subfigure}[b]{0.99\textwidth}
    \centering
    \includegraphics[width=0.85\textwidth]{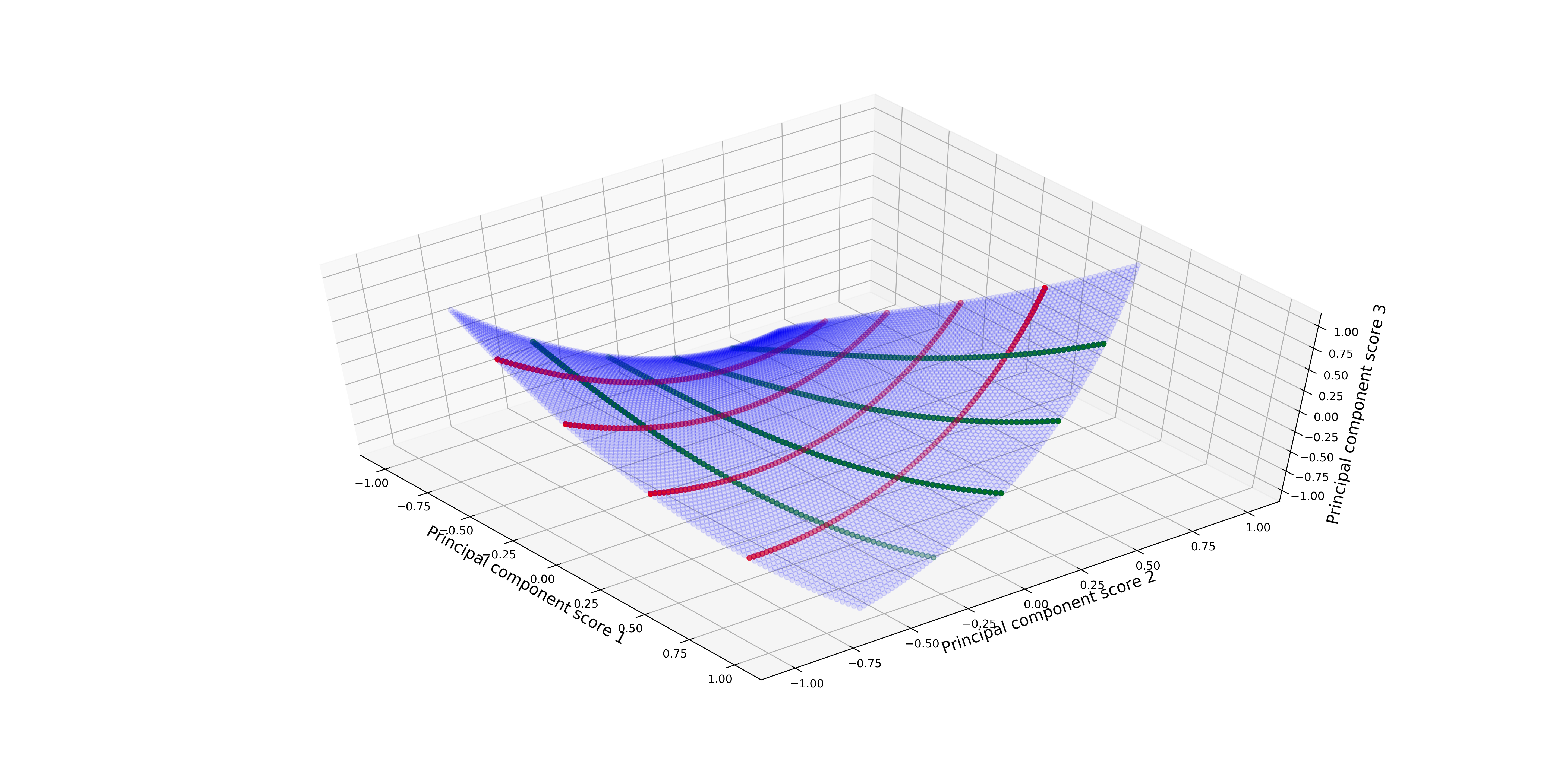}
    \caption{}
    \label{fig:orig_lines}
    \end{subfigure}
    \begin{subfigure}[b]{0.99\textwidth}
    \centering
    \includegraphics[width=0.85\textwidth]{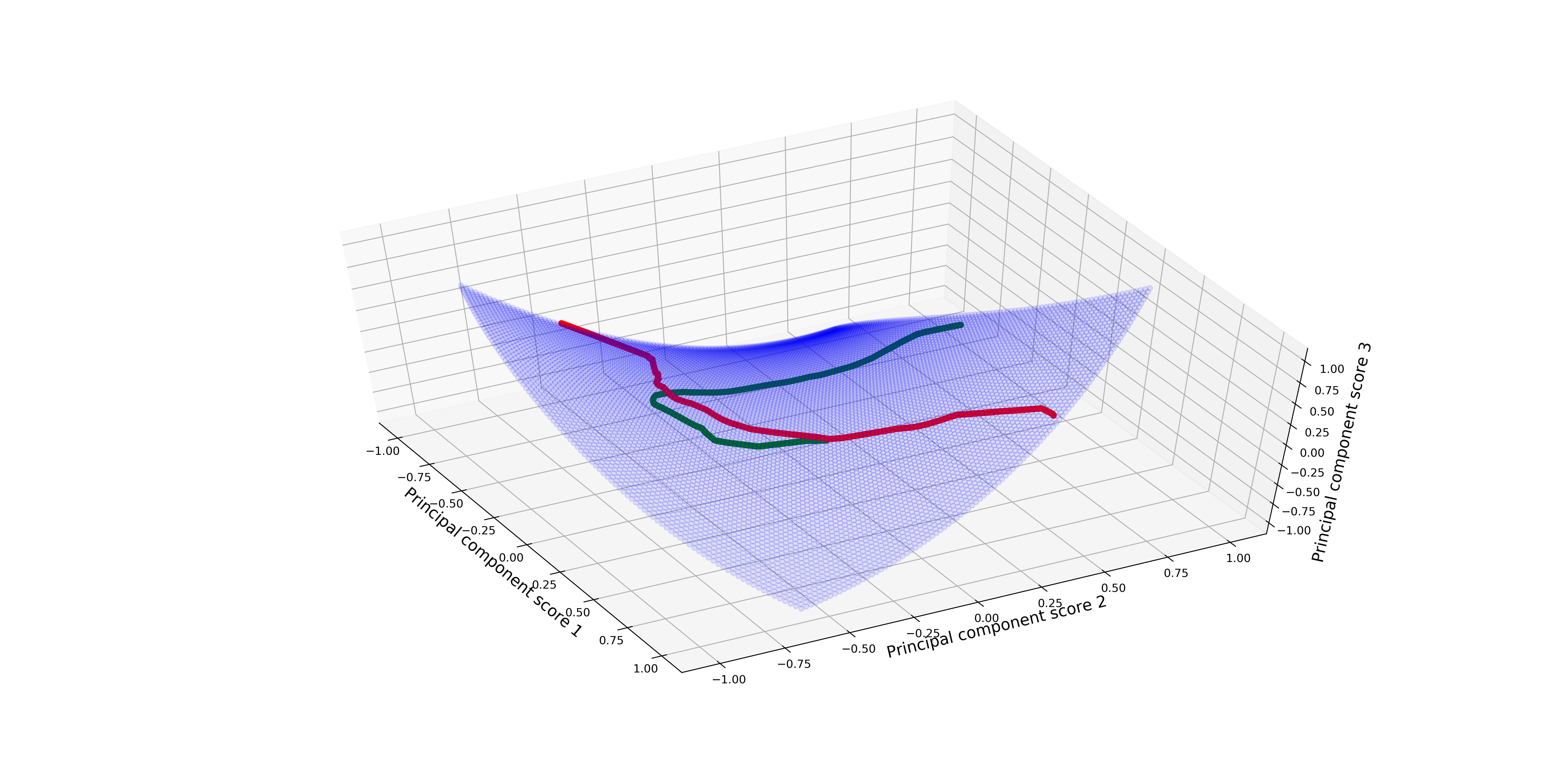}
    \caption{}
    \label{fig:target_entangled}
    \end{subfigure}
    \caption{First three principal component scores of the dataset (blue). On the top, points corresponding to constant damage for spring $1$ and varying damage percentage for spring $2$ (red), and points corresponding to constant damage for spring $2$ and varying damage percentage for spring $1$ (green). In the bottom, GAN-generated samples (red and green) by locking one latent variable to $0$ and varying the other in the interval $[-1, 1]$; regular GAN used.}
    \label{fig:PCA_major_axes_and_samples}
\end{figure}

In order to illustrate the problem of the aforementioned entangled representation, a linear example borrowed from \textit{structural health monitoring} (SHM) is presented. A GAN is used to map a two-dimensional Cartesian space to the manifold formed by collecting \textit{frequency response function} (FRF) samples of a simulated system under various levels of damage. Given a lumped-mass system as an example (Figure \ref{fig:mass_spring_example}), and introducing two damage cases, one for spring $1$ and one for spring $2$, samples of the FRF of the first degree of freedom can be collected (damage is simulated as a stiffness reduction \cite{farrar2012structural}). Performing a PCA on the dataset in order to visualise the data, the first three principal component scores of the dataset are shown in Figures \ref{fig:orig_lines} and \ref{fig:target_entangled} as the blue semiopaque points. The variations applied were stiffness reductions for the first and second spring in the interval $[0\%, 20\%]$ and combinations according to the Cartesian product $[0\%, 20\%] \times [0\%, 20\%]$.

In Figure \ref{fig:orig_lines}, curves are shown, along which one of the damage parameters is constant and the second varies. There are two damage parameters and no other parameter varies during simulation, so the manifold is two-dimensional and each curve is parametrised by one of the two damage parameters.

After collecting the data, they are used to train a GAN. The noise vector is a two-dimensional vector and the latent variables are drawn from uniform distributions over the interval $[-1, 1]$ and the output space is the three-dimensional principal component space of the collected FRFs. As a universal GAN cross-validation scheme does not exist, the GAN was trained according to the procedure followed in the original paper \cite{goodfellow2014generative}. Using the trained generator to output data by varying one latent variable at a time and keeping the other constant, the resulting samples are shown in Figure \ref{fig:target_entangled}. It is clear that a disentangled representation of the manifold has not been achieved, since the green and red lines intersect. The algorithm is not expected to perform so, since there are no restrictions on how the latent variables are used in order to generate artificial data. To achieve a disentangled representation in the current work, the inductive bias of \textit{invertibility} is enforced in the mapping provided by the GAN.

\subsection{Invertibility of the GAN}
\label{sec:invertibility}

In an attempt to avoid entanglement, as shown in Figure \ref{fig:target_entangled}, imposition of the property of invertibility to a GAN is considered in the current work. Given that a mapping is invertible, meaning that it is a bijective mapping from one manifold to another, entanglement in Figure \ref{fig:target_entangled} would probably be alleviated and the mappings should be smoother and more meaningful.

Given a continuous bijection $\phi:M \to N$, where $M$ and $N$ are manifolds, it is straightforward to show that if $c_{1}$, $c_{2}$ are curves in $N$, then $\phi^{-1} \circ c_{1}$ and  $\phi^{-1} \circ c_{2}$ are curves in $M$ with the same number of points of intersection. Considering $\phi: M \to N$ and two curves $c_{1}, c_{2} \in N$ then $\phi^{-1} \circ c_{1} = c'_{1}$ and $\phi^{-1} \circ c_{2} = c'_{2}$ are curves in $M$. For every parameter $t_{i}$ corresponding to the intersections of the two curves $c_{1}$, $c_{2}$, it stands that $c_{1}(t_{i}) = c_{2}(t_{i})$, which leads to $c'_{1}(t_{i}) = c'_{2}(t_{i})$ because $\phi$ is a bijection. Therefore, $c_{1}$ and $c_{2}$ have the same points of intersection as $c'_{1}$ and $c'_{2}$. Consequently, by introducing the invertibility inductive bias in a GAN, helps to avoid mappings like the one shown in Figure \ref{fig:target_entangled}, where two orthonormal axes from the latent variable space are mapped onto the entangled green and red lines. Enforcing invertibility might force the GAN to more efficiently achieve a disentanglement of the features of the data in the, not so rare, case that underlying parameters affect the behaviour of the system in a different way (as in the SHM example presented above) and the mapping from the parameter space to the feature space is a bijection. 

\begin{figure}
    \centering
    \begin{adjustbox}{width=0.9\paperwidth,center}
    \includegraphics[width=1.0\textwidth, height=110pt]{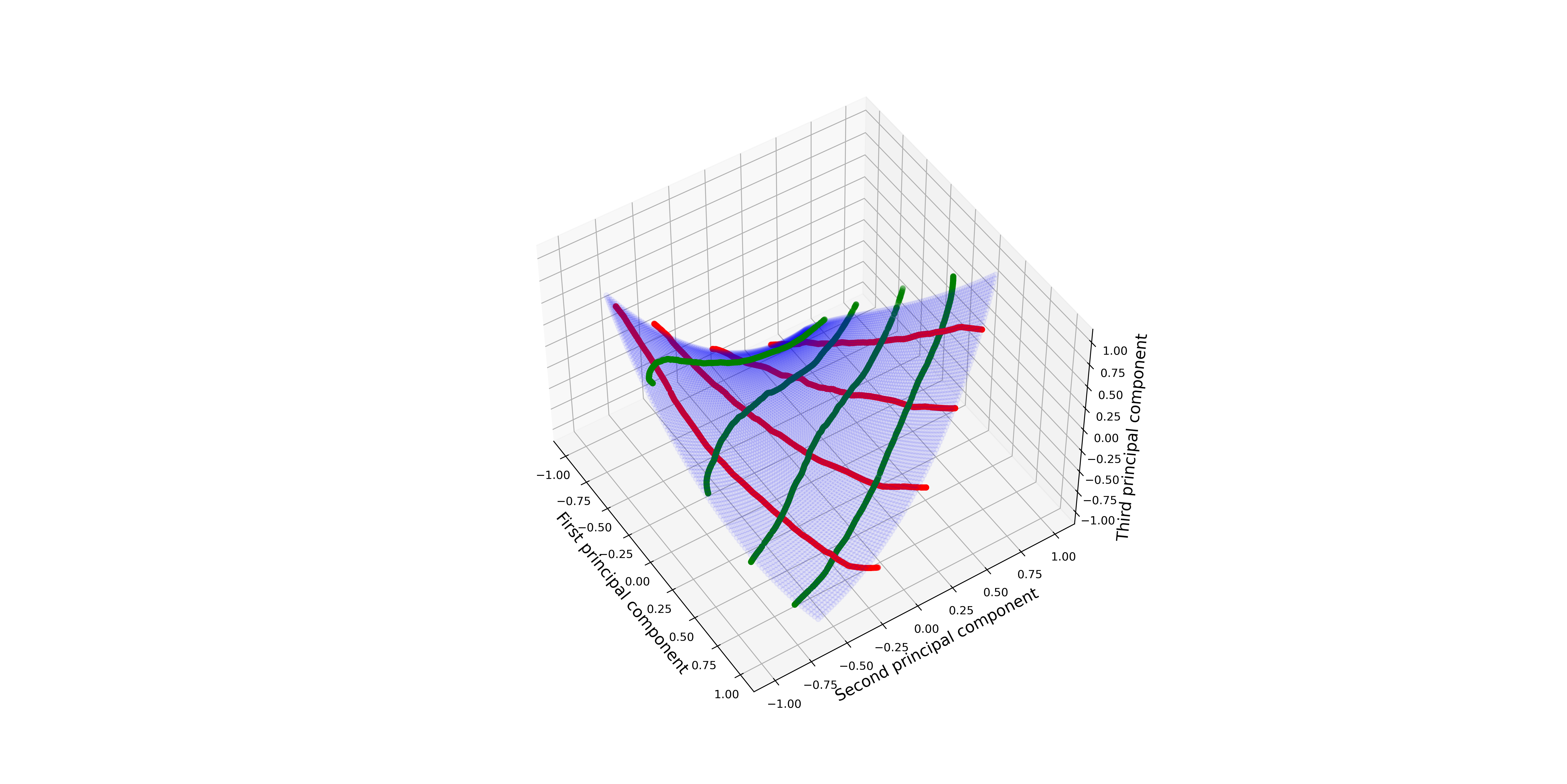}
    \end{adjustbox}
    \caption{First three principal component scores of the dataset (blue) and cycleGAN-generated samples (red and green) by locking one latent variable to various constant values and varying the other the interval $[-1, 1]$.}
    \label{fig:cycleGAN_axes}
\end{figure}

In order to visualise the effects of imposing the property of invertibility on training of the GAN, the cycleGAN algorithm, which will be described in detail, is applied on the same dataset. Similar plots of the results are shown in Figure \ref{fig:cycleGAN_axes}. It is clear that there is no entanglement of features present, in contrast to when training a regular GAN without any restrictions and that the mapping from the two-dimensional noise to the three-dimensional feature space is closer to the mapping shown in Figure \ref{fig:orig_lines}.

For modal analysis, invertibility is quite important. Firstly, it ensures a bijective mapping, and therefore a unique encoding of every different state of the structure into a modal space. Secondly, securing an inverse mapping is fundamental, since a way to perform superposition of the modes is desirable. To satisfy the aforementioned criteria, the algorithm chosen herein is the \textit{cycle-consistent generative adversarial network} \cite{zhu2017unpaired}. In the following sub-sections, this variation of the GAN is presented, together with a way to induce the desired orthogonality of the modes in the mapping.

\subsection{Cycle-GAN}
\label{sec:cycleGAN}
The algorithm chosen to search for invertible mappings under the framework of GANs, is that of the \textit{Cycle-Consistent Adversarial Network} (cycle-GAN) \cite{zhu2017unpaired}. The algorithm was initially introduced as an attempt to transfer images from one style to another; for example, to transform photos into paintings in the style of famous artists. What is interesting, is the way this is achieved; the whole procedure is similar to defining an autoencoder \cite{Kramer1991}, but in terms of a GAN.

A cycle-GAN uses two generators ($G_{X \to Y}$ and $G_{Y \to X}$) and two discriminators ($D_{X}$ and $D_{Y}$). Each pair of networks acts similarly to the classic GAN scheme. The layout of a cycle-GAN is shown in Figures \ref{fig:cycle gan_layout} and \ref{fig:cycle gan_layout_reverse}. The first generator learns to transform samples from domain $X$ into samples of domain $Y$. If used on images, the samples most probably live in the same space, but in different regions or sub-manifolds. The second generator learns to map figures from domain $Y$ back to domain $X$. More specifically, as shown in Figure \ref{fig:cycle gan_layout}, the generator $G_{Y \to X}$ learns to map the sample generated by $G_{X \to Y}$ back to the original sample in domain $X$. Training is performed in two steps; during the first step, the procedure from domain $X$ to domain $Y$ is followed (Figure \ref{fig:cycle gan_layout}) and during the second, the opposite (Figure \ref{fig:cycle gan_layout_reverse}).

\begin{figure}[H]
    \centering
    \begin{adjustbox}{width=0.55\paperwidth,center}
    \begin{tikzpicture}[scale=0.80, every node/.style={transform shape}]
        \definecolor{blue1}{RGB}{0, 128, 255}
        \definecolor{green1}{RGB}{0, 128, 0}
        \node (1) at (0.0, 0.0) [draw, line width=0.5mm, fill=orange, minimum height=1.5cm, minimum width=2cm] {\shortstack{Sample from Domain $X$,\\ $\bm{z_{X}}$}};
        
        \node (2) at (5.0, 0.0) [draw, line width=0.5mm, fill=blue1, minimum height=1.5cm, minimum width=2cm] {Generator $X \to Y$};
        \draw[-{Latex[width=3mm, length=4mm]}, line width=0.5mm] (1) to (2);
        
        \node (3) at (9.8, 0.0) [draw, line width=0.5mm, fill=orange, minimum height=1.5cm, minimum width=2cm] {\shortstack{Sample from Domain $Y$,\\ $\bm{z_{Y}}$}};
        \draw[-{Latex[width=3mm, length=4mm]}, line width=0.5mm] (2) to (3);
        
        \node (4) at (5.0, -3.0) [draw, line width=0.5mm, fill=orange, minimum height=1.5cm, minimum width=2cm] {Domain $Y$ samples};
        
        \node (5) at (9.8, -3.0) [draw, line width=0.5mm, fill=blue1, minimum height=1.5cm, minimum width=2cm] {\shortstack{Discriminator, domain $Y$}};
        
        \node (9) at (9.8, -6.0) [draw, line width=0.5mm, fill=green1, minimum height=1.5cm, minimum width=2cm] {\shortstack{Adversarial Loss}};
        
        \draw[-{Latex[width=3mm, length=4mm]}, line width=0.5mm] (5) to (9);
        
        \draw[-{Latex[width=3mm, length=4mm]}, line width=0.5mm] (3) to (5);
        \draw[-{Latex[width=3mm, length=4mm]}, line width=0.5mm] (4) to (5);
        
        \node (6) at (15.0, 0.0) [draw, line width=0.5mm, fill=blue1, minimum height=1.5cm, minimum width=2cm] {Generator $Y \to X$};
        
        \node (7) at (20.0, 0.0) [draw, line width=0.5mm, fill=orange, minimum height=1.5cm, minimum width=2cm] {\shortstack{Reconstructed sample \\from Domain $X$,\\ $\bm{z_{X}}$}};
        
        \draw[-{Latex[width=3mm, length=4mm]}, line width=0.5mm] (3) to (6);
        \draw[-{Latex[width=3mm, length=4mm]}, line width=0.5mm] (6) to (7);
        
        \node (8) at (20.0, -3.0) [draw, line width=0.5mm, fill=green1, minimum height=1.5cm, minimum width=2cm] {\shortstack{Reconstruction\\loss}};
        
        \draw[-{Latex[width=3mm, length=4mm]}, line width=0.5mm] (7) to (8);
        \draw[-{Latex[width=3mm, length=4mm]}, line width=0.5mm] (1) to[out=290, in=240] (8);
        
    \end{tikzpicture}
    \end{adjustbox}
    \caption{Cycle GAN layout assembled in order to learn the mapping from domain $X$ to domain $Y$ and back. Samples are converted from $X$ to $Y$ by the generator $G_{X \to Y}$. The generated samples are used for adversarial training of the discriminator $D_{Y}$ and the generator $G_{X \to Y}$. Subsequently, the samples are inverse-mapped back to domain $X$ via generator $G_{Y \to X}$ and both generators are trained using the reconstruction-loss error.}
    \label{fig:cycle gan_layout}
\end{figure}
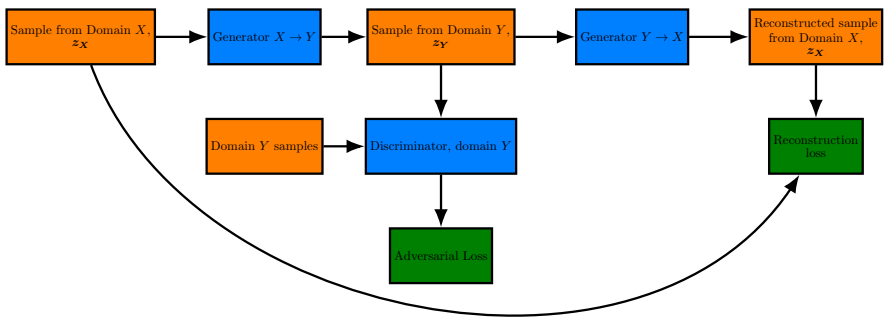

\begin{figure}[H]
    \centering
    \begin{adjustbox}{width=0.55\paperwidth,center}
    \begin{tikzpicture}[scale=0.80, every node/.style={transform shape}]
        \definecolor{blue1}{RGB}{0, 128, 255}
        \definecolor{green1}{RGB}{0, 128, 0}
        \node (1) at (0.0, 0.0) [draw, line width=0.5mm, fill=orange, minimum height=1.5cm, minimum width=2cm] {\shortstack{Sample from Domain $Y$,\\ $\bm{z_{Y}}$}};
        
        \node (2) at (5.0, 0.0) [draw, line width=0.5mm, fill=blue1, minimum height=1.5cm, minimum width=2cm] {Generator $Y \to X$};
        \draw[-{Latex[width=3mm, length=4mm]}, line width=0.5mm] (1) to (2);
        
        \node (3) at (9.8, 0.0) [draw, line width=0.5mm, fill=orange, minimum height=1.5cm, minimum width=2cm] {\shortstack{Sample from Domain $X$,\\ $\bm{z_{X}}$}};
        \draw[-{Latex[width=3mm, length=4mm]}, line width=0.5mm] (2) to (3);
        
        \node (4) at (5.0, -3.0) [draw, line width=0.5mm, fill=orange, minimum height=1.5cm, minimum width=2cm] {Domain $X$ samples};
        
        \node (5) at (9.8, -3.0) [draw, line width=0.5mm, fill=blue1, minimum height=1.5cm, minimum width=2cm] {\shortstack{Discriminator, domain $X$}};
        
        \node (9) at (9.8, -6.0) [draw, line width=0.5mm, fill=green1, minimum height=1.5cm, minimum width=2cm] {\shortstack{Adversarial Loss}};
        
        \draw[-{Latex[width=3mm, length=4mm]}, line width=0.5mm] (5) to (9);
        
        \draw[-{Latex[width=3mm, length=4mm]}, line width=0.5mm] (3) to (5);
        \draw[-{Latex[width=3mm, length=4mm]}, line width=0.5mm] (4) to (5);
        
        \node (6) at (15.0, 0.0) [draw, line width=0.5mm, fill=blue1, minimum height=1.5cm, minimum width=2cm] {Generator $X \to Y$};
        
        \node (7) at (20.0, 0.0) [draw, line width=0.5mm, fill=orange, minimum height=1.5cm, minimum width=2cm] {\shortstack{Reconstructed sample \\from Domain $Y$,\\ $\bm{z_{Y}}$}};
        
        \draw[-{Latex[width=3mm, length=4mm]}, line width=0.5mm] (3) to (6);
        \draw[-{Latex[width=3mm, length=4mm]}, line width=0.5mm] (6) to (7);
        
        \node (8) at (20.0, -3.0) [draw, line width=0.5mm, fill=green1, minimum height=1.5cm, minimum width=2cm] {\shortstack{Reconstruction\\loss}};
        
        \draw[-{Latex[width=3mm, length=4mm]}, line width=0.5mm] (7) to (8);
        \draw[-{Latex[width=3mm, length=4mm]}, line width=0.5mm] (1) to[out=290, in=240] (8);
        
    \end{tikzpicture}
    \end{adjustbox}
    \caption{Cycle GAN layout assembled in order to learn the mapping from domain $Y$ to domain $X$ and backwards. Anti-symmetrical to the procedure shown in Figure \ref{fig:cycle gan_layout}.}
    \label{fig:cycle gan_layout_reverse}
\end{figure}
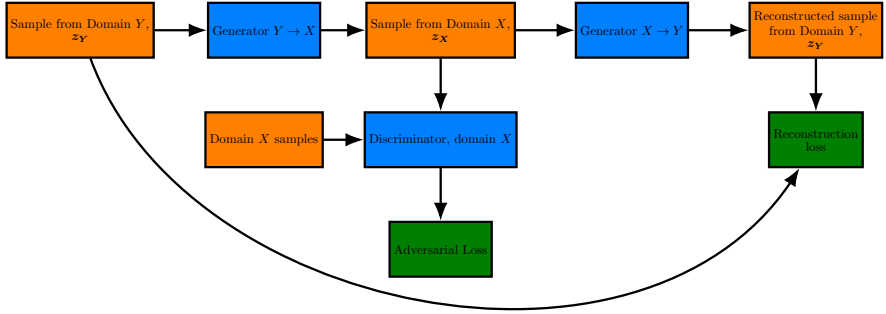

In both steps, the \textit{adversarial loss} $\pazocal{L}_{1}$ is computed, exactly as in the GAN scheme from,
\begin{equation}
    \label{eq:adv_loss}
    \begin{aligned}
    \pazocal{L}_{1} (G_{X \to Y}, D_{Y}, X, Y) = \mathbb{E}_{\bm{y} \sim p_{y}(\bm{y})}[\log D_{Y}(y)] + \\
    \mathbb{E}_{\bm{x} \sim p_{x}(\bm{x})}[\log(1 - D_{Y}(G_{X \to Y}(\bm{x})))]
    \end{aligned}
\end{equation}
and accordingly for the inverse training step. The second type of loss used in training - \textit{the reconstruction loss} (\textit{cycle loss} in the original work) - is given by,
\begin{equation}
    \label{eq:cycle_loss}
    \begin{aligned}
    \pazocal{L}_{2}(G_{X \to Y}, G_{Y \to X}) = \mathbb{E}_{\bm{x} \sim p_{x}(\bm{x})}[\norm{G_{Y \to X}(G_{X \to Y}(\bm{x})) - x}_{n}] + \\ \mathbb{E}_{\bm{y} \sim p_{y}(\bm{y})}[\norm{G_{X \to Y}(G_{Y \to X}(\bm{y})) - y}_{n}]
    \end{aligned}
\end{equation}
where $\norm{\ }_{n}$ is the $n^{th}$ order norm. In the original work, a first-order norm was used, but in the current work, a second-order one yielded better results.

The total training loss is computed from,
\begin{equation}
    \begin{aligned}
    \pazocal{L}(G_{X \to Y}, G_{Y \to X}, D_{X}, D_{Y}) = \pazocal{L}_{1}(G_{X \to Y}, D_{Y}, X, Y) + \\ 
    \pazocal{L}_{1}(G_{Y \to X}, D_{X}, Y, X) + \\ 
    \lambda \pazocal{L}_{2}(G_{X \to Y}, G_{Y \to X})
    \end{aligned}
    \label{eq:total_loss}
\end{equation}
where $\lambda$ controls the relative importance between the adversarial loss and the reconstruction loss (in the original work, a nominal value suggested is $\lambda=10$). The optimisation problem solved is,
\begin{equation}
    \begin{aligned}
    G_{X \to Y}^{*}, G_{Y \to X}^{*} = \min\limits_{G_{X \to Y}, G_{Y \to X}}\max\limits_{D_{X}, D_{Y}} \pazocal{L} (G_{X \to Y}, G_{Y \to X}, D_{X}, D_{Y})
    \end{aligned}
    \label{eq:opt_problem}
\end{equation}

In practice, training is performed in the two stages described by Figures \ref{fig:cycle gan_layout} and \ref{fig:cycle gan_layout_reverse}, and each stage comprises three training steps. The first two steps are similar to the GAN scheme described in Section \ref{sec:vanilla}. The newly-introduced third step is that of the reconstruction loss. Samples that change domain in each stage via each generator, are mapped back to their original domain and the reconstruction loss is computed. The error is back-propagated and both generators' trainable parameters are calibrated according to it.

Using this scheme instead of a \textit{vanilla} GAN ensures the invertibility of the mappings and therefore the advantages discussed in the previous section. The discriminators, for the purposes of the current work, have an auxiliary role in the training procedure. 

\subsection{Orthogonality enforcement}
\label{sec:orthogonality}

Orthogonality constraints have been used in GANs before \cite{muller2019orthogonal}, with a view to dealing with mode (not in a structural modal analysis sense) collapse problems \cite{li110tackling}. Orthogonality may be considered as another inductive bias that users impart in training, in order for the results to be closer to their physical understanding of the data. In some cases, orthogonality may assist in achieving disentanglement of features. In the current work, orthogonality is a desired property of the modal analysis procedure. As described in the next sections, the approach to be followed, using cycle GANs, assumes domain $X$ to be the natural coordinate space of the displacements of some structure and domain $Y$ the modal space. Under this framework, it is desired that samples mapped from modal to natural coordinates be orthogonal when they correspond to different modes, enforcing the orthogonality of the mode shapes.

\begin{figure}[]
    \centering
    \begin{adjustbox}{width=0.7\paperwidth,center}
    \begin{tikzpicture}
        \node[] (2D) at (0.0, 0.0) {\includegraphics[scale=0.3]{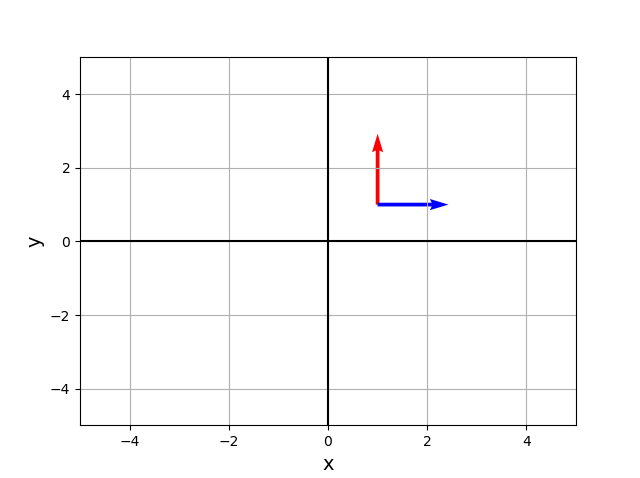}};
        
        \node[] (target) at (8.0, 0.0) {\includegraphics[scale=0.3]{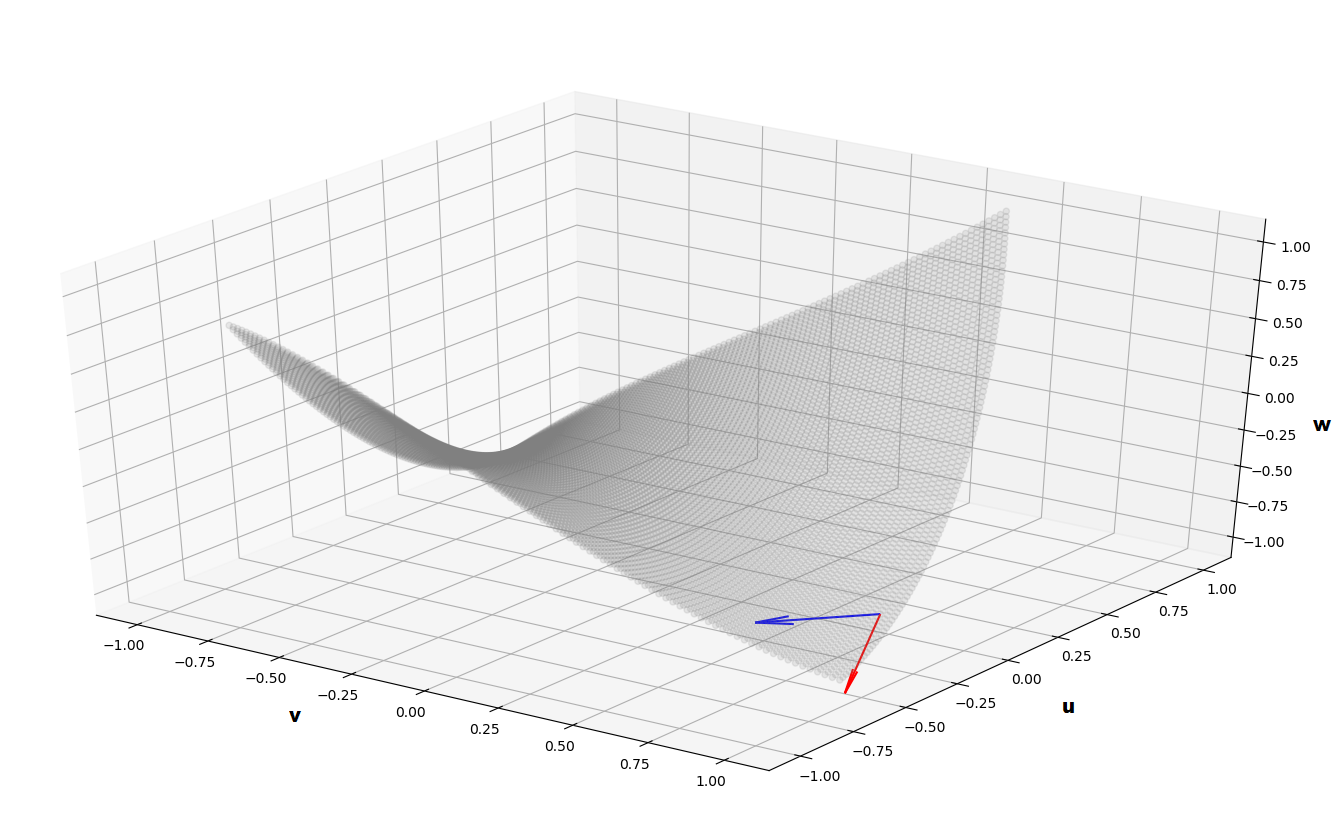}};
        
        \draw[->, thick] (2.0, 1.2) to[out=45,in=135] (5.5, 1.0) node[midway, above] {};
        \node[] (gen) at (3.75, 2.2) {Generator};
    \end{tikzpicture}
    \end{adjustbox}
    \caption{Preservation of orthogonality of vectors by the mapping of the generator from the source Cartesian space ($x, y$ coordinate system) to the target manifold ($u, v, w$ coordinate system).}
    \label{fig:orthogonality_schematically}
\end{figure}

Departing for a while from the modal framework and returning to GANs, orthogonality enforcement means that ``two vectors tangent to the latent manifold and parallel to two of the axes of the latent space are, of course, orthogonal and shall remain orthogonal in the real space. A schematic representation of the idea is shown in Figure \ref{fig:orthogonality_schematically}. Trying to enforce this behaviour in the generator, a new assembly of networks is defined with a view to locally ensuring the orthogonality of the partial derivatives of the mappings from the modal space to the real space. Mappings with such behaviour are called \textit{conformal} or \textit{angle-preserving}.

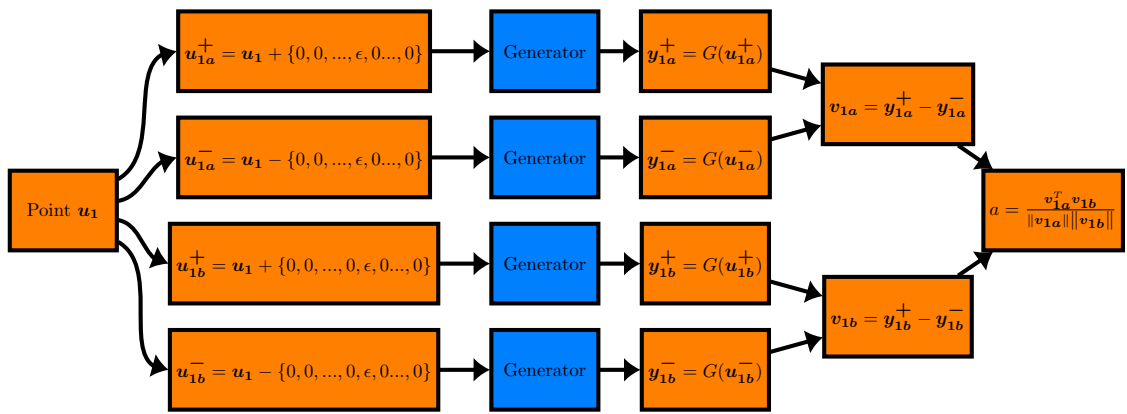
\begin{figure}[]
    \centering
    \begin{adjustbox}{width=0.7\paperwidth,center}
    \begin{tikzpicture}[scale=0.65, every node/.style={transform shape}]
        \definecolor{blue1}{RGB}{0, 128, 255}
        \node (1) at (-0.5, -2.0) [draw, line width=0.5mm, fill=orange, minimum height=1.5cm, minimum width=2cm] {Point $\bm{u_{1}}$};
        \node (2) at (4.0, 1.0) [draw, line width=0.5mm, fill=orange, minimum height=1.5cm, minimum width=2cm] {$\bm{u_{1a}^{+}} = \bm{u_{1}} + \{0, 0, ..., \epsilon, 0..., 0\}$};
        \node (3) at (4.0, -1.0) [draw, line width=0.5mm, fill=orange, minimum height=1.5cm, minimum width=2cm] {$\bm{u_{1a}^{-}} = \bm{u_{1}} - \{0, 0, ..., \epsilon, 0..., 0\}$};
        
        \node (g1) at (8.5, 1.0) [draw, line width=0.5mm, fill=blue1, minimum height=1.5cm, minimum width=2cm] {Generator};
        
        \node (g2) at (8.5, -1.0) [draw, line width=0.5mm, fill=blue1, minimum height=1.5cm, minimum width=2cm] {Generator};
        
        \node (5) at (4.0, -3.0) [draw, line width=0.5mm, fill=orange, minimum height=1.5cm, minimum width=2cm] {$\bm{u_{1b}^{+}} = \bm{u_{1}} + \{0, 0, ..., 0, \epsilon, 0..., 0\}$};
        \node (6) at (4.0, -5.0) [draw, line width=0.5mm, fill=orange, minimum height=1.5cm, minimum width=2cm] {$\bm{u_{1b}^{-}} = \bm{u_{1}} - \{0, 0, ..., 0, \epsilon, 0..., 0\}$};
        
        \node (g3) at (8.5, -3.0) [draw, line width=0.5mm, fill=blue1, minimum height=1.5cm, minimum width=2cm] {Generator};
        
        \node (g4) at (8.5, -5.0) [draw, line width=0.5mm, fill=blue1, minimum height=1.5cm, minimum width=2cm] {Generator};
        
        \node (g11) at (11.5, 1.0) [draw, line width=0.5mm, fill=orange, minimum height=1.5cm, minimum width=2cm] {$\bm{y_{1a}^{+}} = G(\bm{u_{1a}^{+}})$};
        
        \node (g12) at (11.5, -1.0) [draw, line width=0.5mm, fill=orange, minimum height=1.5cm, minimum width=2cm] {$\bm{y_{1a}^{-}} = G(\bm{u_{1a}^{-}})$};
        
        \node (g13) at (11.5, -3.0) [draw, line width=0.5mm, fill=orange, minimum height=1.5cm, minimum width=2cm] {$\bm{y_{1b}^{+}} = G(\bm{u_{1b}^{+}})$};
        
        \node (g14) at (11.5, -5.0) [draw, line width=0.5mm, fill=orange, minimum height=1.5cm, minimum width=2cm] {$\bm{y_{1b}^{-}} = G(\bm{u_{1b}^{-}})$};
        
        \node (v1) at (15.1, -0.0) [draw, line width=0.5mm, fill=orange, minimum height=1.5cm, minimum width=2cm] {$\bm{v_{1a}} = \bm{y_{1a}^{+}} - \bm{y_{1a}^{-}}$};
        
        \node (v2) at (15.1, -4.0) [draw, line width=0.5mm, fill=orange, minimum height=1.5cm, minimum width=2cm] {$\bm{v_{1b}} = \bm{y_{1b}^{+}} - \bm{y_{1b}^{-}}$};
        
        \node (dot) at (18.0, -2.0) [draw, line width=0.5mm, fill=orange, minimum height=1.5cm, minimum width=2cm] {$a = \frac{\bm{v_{1a}}^{T} \bm{v_{1b}}}{\norm{\bm{v_{1a}}} \norm{\bm{v_{1b}}}}$};
        
        \draw[-{Latex[width=3mm, length=2mm]}, line width=0.5mm] (1) to[out=30, in=180] (2);
        \draw[-{Latex[width=3mm, length=2mm]}, line width=0.5mm] (1) to[out=10, in=180] (3);
        \draw[-{Latex[width=3mm, length=2mm]}, line width=0.5mm] (1) to[out=-10, in=180] (5);
        \draw[-{Latex[width=3mm, length=2mm]}, line width=0.5mm] (1) to[out=-30, in=180] (6);
        
        \draw[-{Latex[width=3mm, length=2mm]}, line width=0.5mm] (2) to (g1);
        \draw[-{Latex[width=3mm, length=2mm]}, line width=0.5mm] (3) to (g2);
        \draw[-{Latex[width=3mm, length=2mm]}, line width=0.5mm] (5) to (g3);
        \draw[-{Latex[width=3mm, length=2mm]}, line width=0.5mm] (6) to (g4);
        
        \draw[-{Latex[width=3mm, length=2mm]}, line width=0.5mm] (g1) to (g11);
        \draw[-{Latex[width=3mm, length=2mm]}, line width=0.5mm] (g2) to (g12);
        \draw[-{Latex[width=3mm, length=2mm]}, line width=0.5mm] (g3) to (g13);
        \draw[-{Latex[width=3mm, length=2mm]}, line width=0.5mm] (g4) to (g14);
        
        \draw[-{Latex[width=3mm, length=2mm]}, line width=0.5mm] (g11) to (v1);
        \draw[-{Latex[width=3mm, length=2mm]}, line width=0.5mm] (g12) to (v1);
        
        \draw[-{Latex[width=3mm, length=2mm]}, line width=0.5mm] (g13) to (v2);
        \draw[-{Latex[width=3mm, length=2mm]}, line width=0.5mm] (g14) to (v2);
        
        \draw[-{Latex[width=3mm, length=2mm]}, line width=0.5mm] (v1) to (dot);
        \draw[-{Latex[width=3mm, length=2mm]}, line width=0.5mm] (v2) to (dot);
        
    \end{tikzpicture}
    \end{adjustbox}
    \caption{Orthogonality enforcement assembly using the generator that maps samples from the source Cartesian space to the target manifold.}
    \label{fig:orthogonality_assembly_layout}
\end{figure}

The layout of the assembly used to enforce orthogonality is schematically shown in Figure \ref{fig:orthogonality_assembly_layout}. The goal is to maintain orthogonality of the grid of the latent space into the real space. As shown in Figure \ref{fig:orthogonality_assembly_layout}, a random latent point $\bm{u_{1}}$ is sampled. An axis/variable in the latent space is then chosen, and a small quantity $\epsilon$ is added to that latent coordinate to get the point $\bm{u_{1a}^{+}} = \bm{u_{1}} + \{0, 0, ..., \epsilon, 0..., 0\}$. In the same way, but by subtracting $\epsilon$, yields the point $\bm{u_{1a}^{-}} = \bm{u_{1}} - \{0, 0, ..., \epsilon, 0..., 0\}$. Afterwards, another axis/latent variable is chosen and the same procedure is repeated generating points $\bm{u_{1b}^{+}} = \bm{u_{1}} + \{0, 0, ..., 0, \epsilon, 0..., 0\}$ and $\bm{u_{1b}^{-}} = \bm{u_{1}} - \{0, 0, ..., 0, \epsilon, 0..., 0\}$.

All these points are passed through the generator and their real space counterparts are generated ($\bm{y_{1a}^{+}} = G(\bm{u_{1a}^{+}})$, $\bm{y_{1a}^{-}} = G(\bm{u_{1a}^{-}})$, $\bm{y_{1b}^{+}} = G(\bm{u_{1b}^{+}})$ and $\bm{y_{1b}^{-}} = G(\bm{u_{1b}^{-}})$). Consequently, the vectors $\bm{v_{1a}}$ and $\bm{v_{1b}}$ are computed as $\bm{v_{1a}} = \bm{y_{1a}^{+}} - \bm{y_{1a}^{-}}$ and $\bm{v_{1b}} = \bm{y_{1b}^{+}} - \bm{y_{1b}^{-}}$. Finally, the inner product between $\bm{v_{1a}}$ and $\bm{v_{1b}}$ is computed and divided by the quantity $\norm{\bm{v_{1a}}} \norm{\bm{v_{1b}}}$ to get the cosine of the angle between the two vectors. Optimising the Generator's weights so that this quantity is as close to zero as possible will enforce the desired orthogonality. The orthogonality assembly proposed is in fact calculating the numerical gradient (in geometrical terms $\frac{\partial}{\partial u_{i}}$). along two different axes of a specific point in the manifold in the real space. In the case of modal analysis, as it will be described, if this bias is enforced in the generator mapping from modal to natural coordinates, the orthogonality of the mode shapes is ensured.

\section{Performing nonlinear modal analysis using a cycle-GAN}
\label{sec:ModalCycle}

As described, the cycle-GAN provides a practical method for creating bijective mappings from a source domain to a target domain; this is useful for modal analysis. Given a linear structure, mode shapes provide a representative way of analysing the various independent ways (\textit{modes}) that the structure oscillates under some external load. The modes in such cases, can be used to decompose any response of the structure into independent responses, each one referring to a different natural frequency. Unfortunately this is not the case for nonlinear structures.

In linear structures, modal analysis can be performed either by eigenvalue analysis of the structural parameters (mass and stiffness matrices), or in an operational manner, via \textit{principal component analysis} PCA \cite{wold1987principal} of the displacements (or accelerations) given by sensors placed on a structure; this procedure in some cases coincides with linear modal analysis. For structures with nonlinearities, these methods cannot be applied. PCA itself is a linear method and therefore provides only a linear decomposition of the data. Motivated by PCA's data-driven scheme, an attempt to perform similarly data-driven nonlinear modal analysis was proposed in \cite{worden2017machine} and \cite{dervilis2019nonlinear}. 

The decomposition in \cite{worden2017machine} aimed at maintaining the statistical independence and orthogonality aspects of a modal analysis. More specifically, a genetic algorithm was used to learn a decomposition of the displacements into latent variables with correlation close to $0$. The approach performed better than a linear PCA analysis of the data and, under the criteria of \cite{worden2017machine, dervilis2019nonlinear}, in the case of a two-degree-of-freedom lumped-mass system with a cubic nonlinearity, a very efficient decomposition into modes was achieved. However, as the degrees-of-freedom of the systems increase, the algorithm seems to not perform equally well, because the mapping was more complicated.

A major drawback of the algorithm is that the maps used in order to perform the decomposition are restricted to a fixed polynomial order. For example, for a three-degree-of-freedom system, the equation used to transform the physical coordinates $\{y_{1}, y_{2}, y_{3}\}^{T}$ into the corresponding modal $\{u_{1}, u_{2}, u_{3}\}^{T}$ was,
\begin{equation}
    \begin{Bmatrix} u_{1} \\ u_{2} \\ u_{3} \end{Bmatrix} = \begin{bmatrix} a_{11} & a_{12} & a_{13} \\ a_{21} & a_{22} & a_{23} \\ a_{31} & a_{32} & a_{33} \end{bmatrix} \begin{Bmatrix} y_{1} \\ y_{2} \\ y_{3} \end{Bmatrix} + \begin{bmatrix} b_{11} & \dots & a_{19} \\ b_{21} & \dots & b_{29} \\ b_{31} & \dots & a_{39} \end{bmatrix}\begin{Bmatrix} y_{1}^{3} \\ y_{1}^{2}y_{2} \\ y_{1}^{2}y_{3} \\ y_{2}^{3} \\ y_{2}^{2}y_{1} \\ y_{2}^{2}y_{3} \\ y_{3}^{2} \\ y_{3}^{2}y_{1} \\ y_{3}^{2}y_{2}\end{Bmatrix}
    \label{eq:decomposition_equation}
\end{equation}
where $a_{ij}, i, j \in \{1, 2, 3\}$ and $b_{i,j}, i\in \{1, 2, 3\}, j \in \{1, 2..., 9\}$ are the tunable parameters that are optimised using the objective function,
\begin{equation}
    \begin{aligned}
    J &= |\{A_{1}\} \cdot \{A_{2}\}| + |\{A_{1}\} \cdot \{A_{3}\}| + |\{A_{2}\} \cdot \{A_{3}\}| \\
    & + \Cor(u_1, u_2) + \Cor(u_1, u_3) + \Cor(u_2, u_3) \\
    & + \Cor(u_1^3, u_2) + \Cor(u_1^3, u_3) + \Cor(u_2^3, u_1) \\
    & + \Cor(u_2^3, u_3) + \Cor(u_3^3, u_1) + \Cor(u_3^3, u_2)
    \end{aligned}
    \label{eq:obj_function}
\end{equation}
where $A_{i} = \{a_{1i}, a_{2i}, a_{3i}\}$. The problem with the decomposition of equation (\ref{eq:decomposition_equation}) is that it is a truncated polynomial, which is not a universal approximator.

The above objective function, aims at the orthogonality of the modal coordinates and their statistical independence in the $u$ space. In the current work a different approach is followed. Instead of decomposing the dataset of displacements into a modal space, a mapping from a pre-defined latent space to the observed dataset and back will be sought. Using the cycle-GAN scheme, and considering domain $X$ as the displacement domain and domain $Y$ as the modal domain, this composition is attempted. For the rest of the paper, in order to use the same symbols as in \cite{worden2017machine}, the physical domain and its corresponding coordinates will be $Y$ and $\bm{y}_{i}$ respectively and the `modal space' and coordinates will be $U$ and $\bm{u}_{i}$ respectively.

For the purposes of modal analysis and trying to impose the statistical independence of the modal coordinates $\bm{u}_{i}$, they are chosen to be $n$-dimensional random Gaussian vectors, with mean values equal to zero and correlation matrix equal to $I_n$ which is an $n$-dimensional identity matrix. Using this modal space, all correlation terms described as in equation (\ref{eq:obj_function}) are expected to be zero.

To help ensure the statistical independence of the modal coordinates, PCA was applied to the physical coordinates before training the cycle-GAN. PCA is selected in order to minimise the initial correlations and because it is a linear and easily-invertible transformation.

In addition to minimising linear correlations between modal coordinates, PCA is expected to assist further. Assuming that the system's response can be decomposed in modes which are excited on a different level and contribute unevenly to the total movement, a scaling problem arises. The axes of the modal space correspond to curves in the physical space. These curves, in the linear case, are the (linear) axes of the ellipsoid, which is formed if one plots points that correspond to displacements of the structure's degrees-of-freedom in the physical displacement space. By introducing nonlinearities to the system, the aforementioned ellipsoid is deformed and its axes are no longer lines, they become curves. The algorithm proposed is called to find a mapping $\phi$, which maps the axes of the predefined modal coordinates onto the deformed-by-the-nonlinearity axes of the aforementioned ellipsoid. Because every mode contributes on a different level to the total motion of the structure, some of these axes are longer than others. By performing PCA, the coordinate system is rotated so that the axes' variance are in descending order. Subsequently, by scaling every coordinate of the samples in the interval $[-1, 1]$, as needed for the neural networks to perform, it is expected that the axes representing each mode will be scaled almost similarly, facilitating the task of the cycle-GAN; i.e.\ defining a mapping from the modal coordinates to the physical coordinates.

By optimising with a target value of zero for the cosine computed in the rightmost block of computation in Figure \ref{fig:orthogonality_assembly_layout}, distinct modal coordinates are forced to generate samples in the physical space that are orthogonal to the ones generated by different mode variables; this way imposing the orthogonality of the mode shapes and maintaining locally the orthogonality of the aforementioned curved axes in the real space. The major advantage of the proposed cycle-GAN approach is that, since the neural networks concerned can approximate any function \cite{csaji2001approximation}, it is not restricted by the order of the terms used in equation (\ref{eq:decomposition_equation}). Furthermore, the inverse mapping provides the nonlinear superposition mapping required for the sake of completeness of the method. In the next sections, applications of the algorithm on simulated and experimental data are presented.

\section{Case studies}
\label{sec:case_studies}

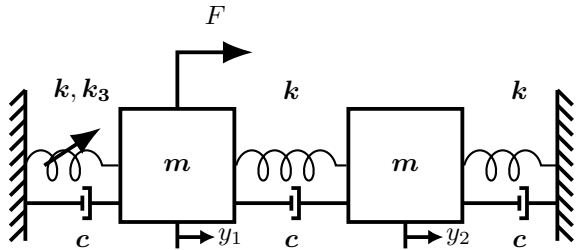
\begin{figure}[h!]
    \centering
    \begin{tikzpicture}
        \draw[line width=0.5mm] (-2,1) -- (-2,-1);

        \draw[line width=0.5mm] (-2, 1) -- (-2.2, 0.8);
        \draw[line width=0.5mm] (-2, 0.8) -- (-2.2, 0.6);
        \draw[line width=0.5mm] (-2, 0.6) -- (-2.2, 0.4);
        \draw[line width=0.5mm] (-2, 0.4) -- (-2.2, 0.2);
        \draw[line width=0.5mm] (-2, 0.2) -- (-2.2, 0.0);
        \draw[line width=0.5mm] (-2, 0.0) -- (-2.2, -.2);
        \draw[line width=0.5mm] (-2, -.2) -- (-2.2, -.4);
        \draw[line width=0.5mm] (-2, -.4) -- (-2.2, -.6);
        \draw[line width=0.5mm] (-2, -.6) -- (-2.2, -.8);
        \draw[line width=0.5mm] (-2, -.8) -- (-2.2, -1.0);
    
        \node (1) at (0.0, 0.0) [draw, line width=0.5mm, minimum width=1.5cm, minimum height=1.5cm] {$\bm{m}$};
        \node (2) at (3.0, 0) [draw, line width=0.5mm, minimum width=1.5cm, minimum height=1.5cm] {$\bm{m}$};
        
        \draw[thick, decoration={aspect=0.65, segment length=3mm,
             amplitude=0.2cm, coil}, decorate] (-2, 0) -- (1);
        
        \draw[-{Latex[width=3mm, length=4mm]}, line width=0.5mm] (-1.75, 0.0) -- (-1.0,  .5);
        
        \draw[thick, decoration={aspect=0.65, segment length=3mm,
             amplitude=0.2cm, coil}, decorate] (1) --(2);
             
         \draw[thick, decoration={aspect=0.65, segment length=3mm,
         amplitude=0.2cm, coil}, decorate] (2) --(5.0, 0);
        
        \draw[line width=0.5mm] (5.0,1) -- (5.0,-1);
        \draw[line width=0.5mm] (5.0, 1) -- (5.2, 0.8);
        \draw[line width=0.5mm] (5.0, 0.8) -- (5.2, 0.6);
        \draw[line width=0.5mm] (5.0, 0.6) -- (5.2, 0.4);
        \draw[line width=0.5mm] (5.0, 0.4) -- (5.2, 0.2);
        \draw[line width=0.5mm] (5.0, 0.2) -- (5.2, 0.0);
        \draw[line width=0.5mm] (5.0, 0.0) -- (5.2, -.2);
        \draw[line width=0.5mm] (5.0, -.2) -- (5.2, -.4);
        \draw[line width=0.5mm] (5.0, -.4) -- (5.2, -.6);
        \draw[line width=0.5mm] (5.0, -.6) -- (5.2, -.8);
        \draw[line width=0.5mm] (5.0, -.8) -- (5.2, -1.0);

        \draw[-{Latex[width=3mm, length=4mm]}, line width=0.5mm] (0, 0.75) -- (0, 1.5) -- (1, 1.5);
        \node[] at (0.5, 2.0) {$F$};
        
        \draw[line width=0.5mm] (0, -0.75) -- (0, -1.1);
        \draw[-{Latex[width=2mm, length=2mm]}, line width=0.5mm] (0, -0.95) -- (0.5, -0.95);
        \node[] at (0.7, -0.95) {$y_{1}$};
        
        \draw[line width=0.5mm] (3.0, -0.75) -- (3.0, -1.1);
        \draw[-{Latex[width=2mm, length=2mm]}, line width=0.5mm] (3.0, -0.95) -- (3.5, -0.95);
        \node[] at (3.7, -0.95) {$y_{2}$};
        
        \node[] at (-1.25, 1.0) {$\bm{k}, \bm{k_{3}}$};
        \node[] at (1.5, 1.0) {$\bm{k}$};
        \node[] at (4.5, 1.0) {$\bm{k}$};
        
        \draw[line width=0.5mm] (-2, -0.5) -- (-1.25, -0.5);
        \draw[line width=0.5mm] (-1.25, -0.4) -- (-1.25, -0.6);
        \draw[line width=0.5mm] (-1.25, -0.3) -- (-1.15, -0.3) -- (-1.15, -0.7) -- (-1.25, -0.7);
        \draw[line width=0.5mm] (-1.15, -0.5) -- (-0.75, -0.5);
        
        \draw[line width=0.5mm] (0.75, -0.5) -- (1.5, -0.5);
        \draw[line width=0.5mm] (1.5, -0.4) -- (1.5, -0.6);
        \draw[line width=0.5mm] (1.5, -0.3) -- (1.6, -0.3) -- (1.6, -0.7) -- (1.5, -0.7);
        \draw[line width=0.5mm] (1.6, -0.5) -- (2.25, -0.5);
        
        \draw[line width=0.5mm] (3.75, -0.5) -- (4.5, -0.5);
        \draw[line width=0.5mm] (4.5, -0.4) -- (4.5, -0.6);
        \draw[line width=0.5mm] (4.5, -0.3) -- (4.6, -0.3) -- (4.6, -0.7) -- (4.5, -0.7);
        \draw[line width=0.5mm] (4.6, -0.5) -- (5.0, -0.5);

        \node[] at (-1.25, -1.0) {$\bm{c}$};
        \node[] at (1.5, -1.0) {$\bm{c}$};
        \node[] at (4.5, -1.0) {$\bm{c}$};

    \end{tikzpicture} 
    \caption{Two degree-of-freedom mass-spring system.}
    \label{fig:mass_spring}
\end{figure}

As in \cite{worden2017machine}, a two-degree-of-freedom system was studied first. In every simulated case study here, the physical system is a lumped-mass system as shown in Figure \ref{fig:mass_spring}. The parameters of the model are the same as in \cite{worden2017machine}, i.e. $m=1.0$, $c=0.1$, $k=10$, $k_{3} = 1500$. The equations of motion of the system are,
\begin{equation}
    \begin{bmatrix} m & 0 \\ 0 & m \end{bmatrix} \begin{Bmatrix} \ddot{y}_{1} \\ \ddot{y}_{2} \end{Bmatrix} + 
    \begin{bmatrix} 2c & -c \\ -c & 2c \end{bmatrix}\begin{Bmatrix} \dot{y}_{1} \\ \dot{y}_{2} \end{Bmatrix} + 
    \begin{bmatrix} 2k & -k \\ -k & 2k \end{bmatrix}\begin{Bmatrix} y_{1} \\ y_{2} \end{Bmatrix} + 
    \begin{Bmatrix} k_{3}y^{3}_{1} \\ 0 \end{Bmatrix} = \begin{Bmatrix} F \\ 0 \end{Bmatrix} 
\end{equation}
The excitation was Gaussian white noise with zero mean and standard deviation $5.0$, low-pass filtered onto the frequency interval $[0, 50]$ Hz. For every system, two datasets of $100000$ points were generated using differently-seeded random excitations. The first dataset was used for training and model selection and the second for testing the efficiency of the algorithm, as well as for the \textit{power spectral density} (PSD) functions presented. The PSDs were calculated using Welch's method \cite{welch1967use}. 

For every case study, the generators used for the cycle-GAN were three-layered neural networks, since they are proven universal approximators \cite{tarassenko1998guide}. All tested neural networks had an input and an output layer, both with neurons equal to the dimension of the problem and a hidden layer whose size was optimised by training networks with different sizes and picking the best according to an inner product criterion described in the next paragraphs. The hidden layer sizes used belonged to the set $\{50, 60, ... 190, 200\}$. For every size, 20 random initialisations of the neural networks were performed. The large number of initialisations is determined by the obvious sensitivity of the algorithm to the initial placements of the principal axes on the target manifold. Moreover, for every generator hidden layer size, the corresponding discriminator had the same hidden layer size. This was done in order to allow both networks to act on the same latent space and in order to reduce the hyperparameters by inducing such a symmetry to the algorithm. The activation functions that performed best were \textit{hyperbolic tangent} for the hidden layer and linear for the output layer, which is actually the best practice for regression problems. As proposed in \cite{zhu2017unpaired}, the value of the $\lambda$ parameter from equation (\ref{eq:total_loss}) is set to $10$ and the training algorithm was the Adam optimiser \cite{kingma2014adam}.

In each case study, the best model was selected as the one that minimised the average cosine between every combination of PSD vectors in the latent space ($(PSD_1, PSD_2)$ for the two-dimensional case, $(PSD_1, PSD_2)$, $(PSD_1, PSD_3)$, $(PSD_2, PSD_3)$ for the three-dimesional, etc.). The equation describing the model selection criterion is defined by,
\begin{equation}
\label{eq:model_selec_eq}
    \pazocal{L}_{\cos} = \sum_{i=1, j=i+1}^{n_{dof}} \frac{PSD_{i} \cdot PSD_{j}}{\norm{PSD_{i}}\norm{PSD_{j}}}
\end{equation}
where $n_{dof}$ is the number of degrees of freedom of the system. The cosine criterion is an attempt to imitate the `by eye' way of choosing the best model based on the isolation of modes in the PSDs. It is not identical, since minimisation of the dot product could be achieved if every PSD was equal to zero for every frequency, except for one; however, it is expected that this will not be the case, because of the orthogonality restriction applied. Model efficiency testing was performed every $100$ epochs of training. The quantity in equation (\ref{eq:model_selec_eq}) cannot go to zero, since damping forces some of the energy in the PSD to concentrate in frequencies around the natural frequencies; nevertheless, it is a convenient measure of separation of the modes.

Throughout the rest of the paper, in order to keep a similar format to \cite{worden2017machine}, PSDs referring to physical coordinates will be drawn with a blue colour, PSDs referring to modal components (or PCA components for the case of the experimental data) will be drawn using a black colour and PSDs referring to latent components of the cycle-GAN approach, presented herein, will be drawn in a red colour.

\subsection{Two-degree-of-freedom system}
\label{sec:2DOF}

\begin{figure}[]
    \begin{adjustbox}{height=160pt, width=0.65\paperwidth,center}
    \includegraphics{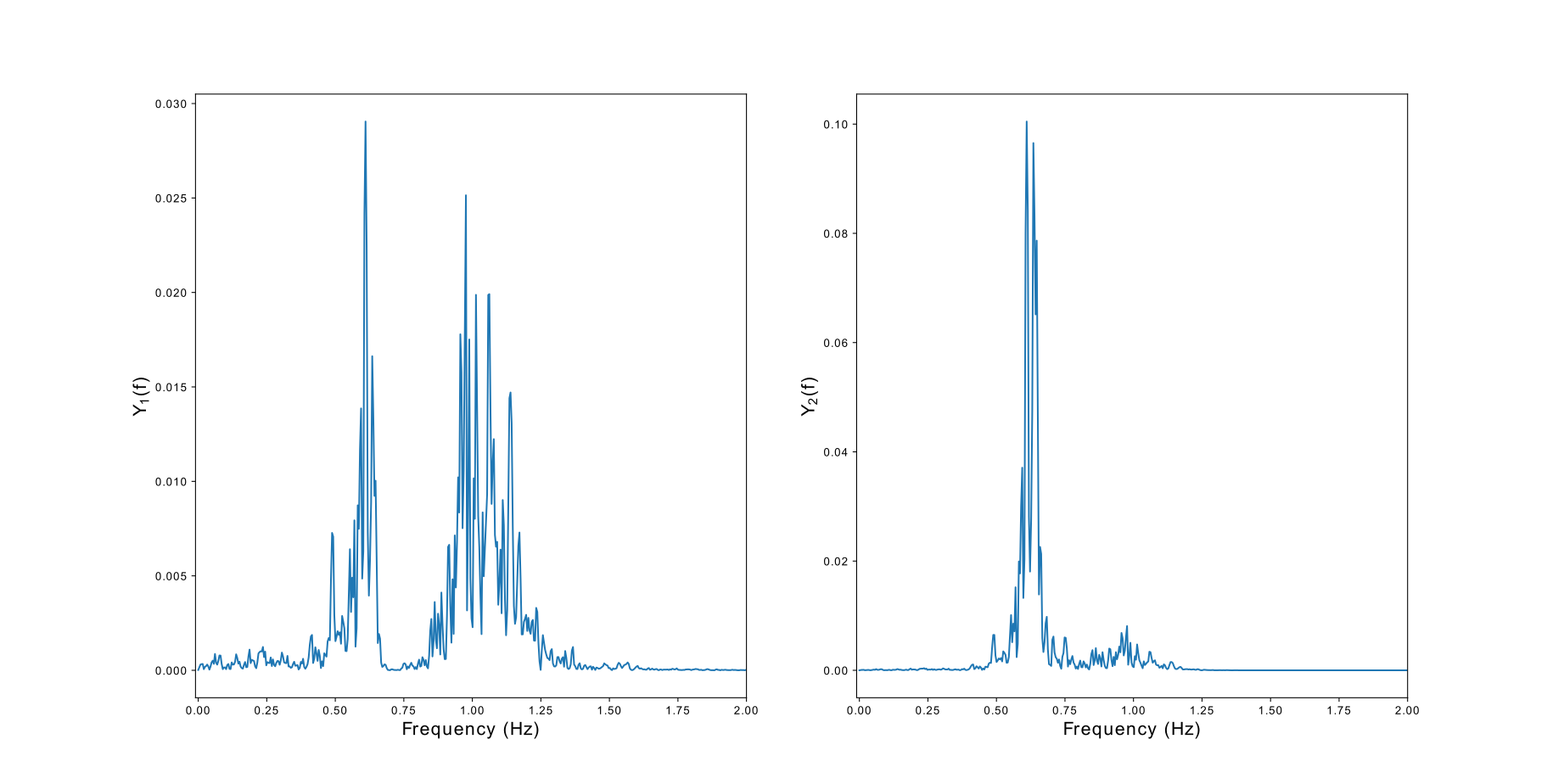}
    \end{adjustbox}
    \caption{PSDs of two-degree-of-freedom structure; physical coordinates.}
    \label{fig:physical_2DOF_PSD}
\end{figure}

The described procedure was initially followed for a two-degree-of-freedom simulated lumped mass system with a cubic nonlinearity (Figure \ref{fig:mass_spring}). The PSDs of the physical degrees of freedom are shown in Figure \ref{fig:physical_2DOF_PSD}. The nonlinearity is clearly affecting the $PSD_1$. A spreading towards higher frequencies is clear in the second mode. Linear natural frequencies are the ones at which the structure absorbs the most energy and are proportional to the stiffness of the structural members. As the nonlinear member in this case is hardening, there are time instants during its movement that its stiffness is higher because of the higher value of displacement of the first degree of freedom, leading to the natural frequencies and energy in the PSD being spread towards higher frequencies. The movement in this case is clear if one considers the natural frequencies of the underlying linear problem, which are $0.5$ Hz and $0.87$ Hz.

\begin{figure}[]
    \begin{adjustbox}{width=0.8\paperwidth, center}
    \includegraphics{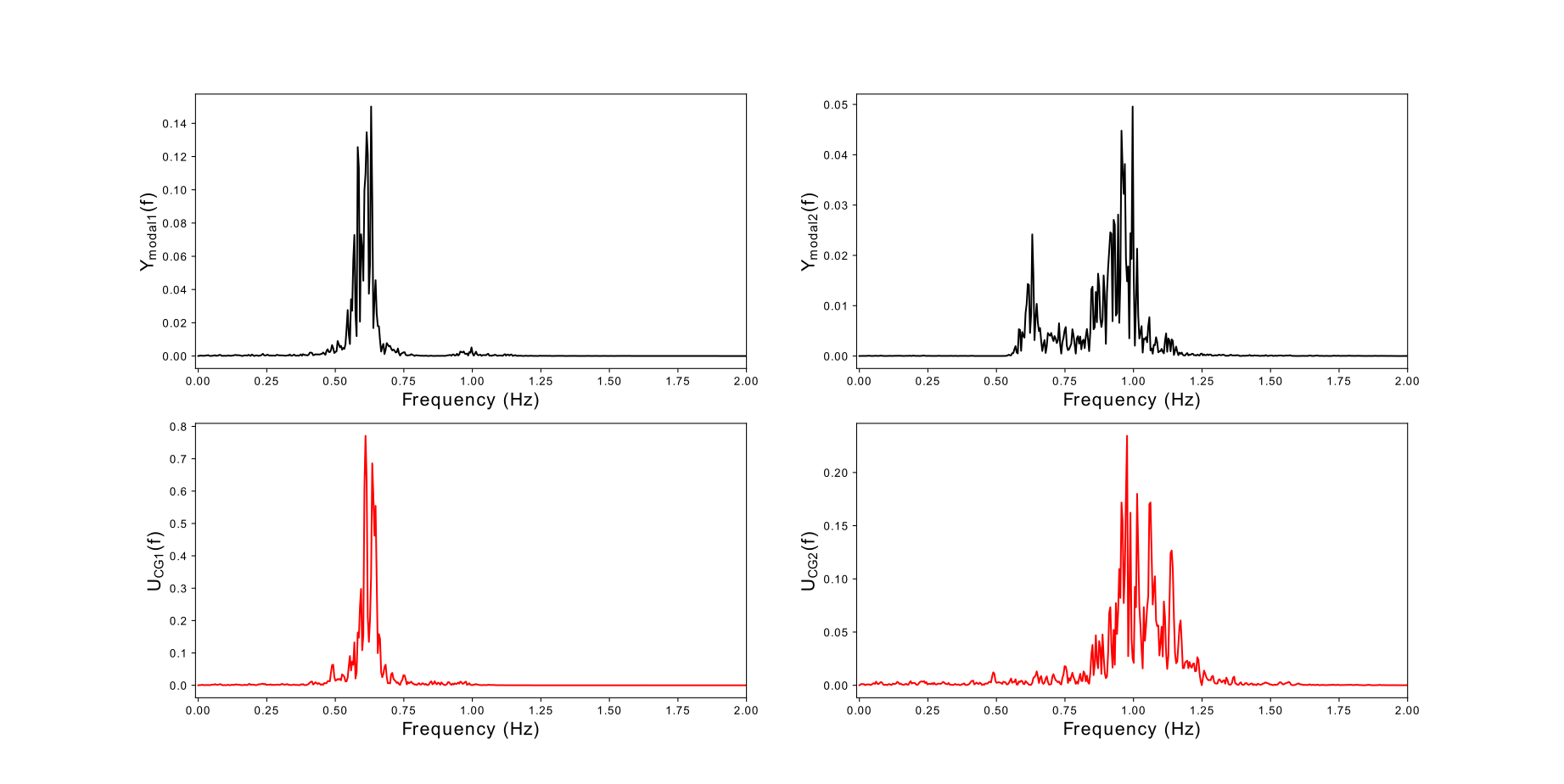}
    \end{adjustbox}
    \caption{PSDs of two-degree-of-freedom structure, linear modal decomposition (top) and cycle-GAN decomposition selected via the inner product criterion (bottom).}
    \label{fig:PCA_CG_2DOF_PSD}
\end{figure}

Using the cycle-GAN to decompose the displacements into modal coordinates, the model that yielded the best results had $100$ units in its hidden layer. The effect of the decomposition performed by that model is shown in the bottom PSDs in Figure \ref{fig:PCA_CG_2DOF_PSD}. For comparison, the PCA decomposition, which in this case coincides with linear modal analysis, is shown on the top row of the same figure. As expected, linear modal analysis cannot decouple the modes, because of the nonlinearity. In contrast, the cycle-GAN algorithm is able to efficiently do that. The result is very similar to that using the SADE algorithm in \cite{worden2017machine}.

The results of this case study, which can be considered a benchmark, are encouraging. The large size of the hidden layer might be a result of selecting the best model, having as the single criterion, the inner product in equation (\ref{eq:model_selec_eq}). Smaller networks might also perform well. If one is concerned about the complexity of the model, a second criterion can be considered along with the inner product for the procedure of model selection. The algorithm is subsequently tested on larger systems and also on an experimental system.

\subsection{Three-degree-of-freedom system}
\label{sec:3DOF}

\begin{figure}[H]
    \begin{adjustbox}{height=140pt, width=0.75\paperwidth, center}
    \includegraphics{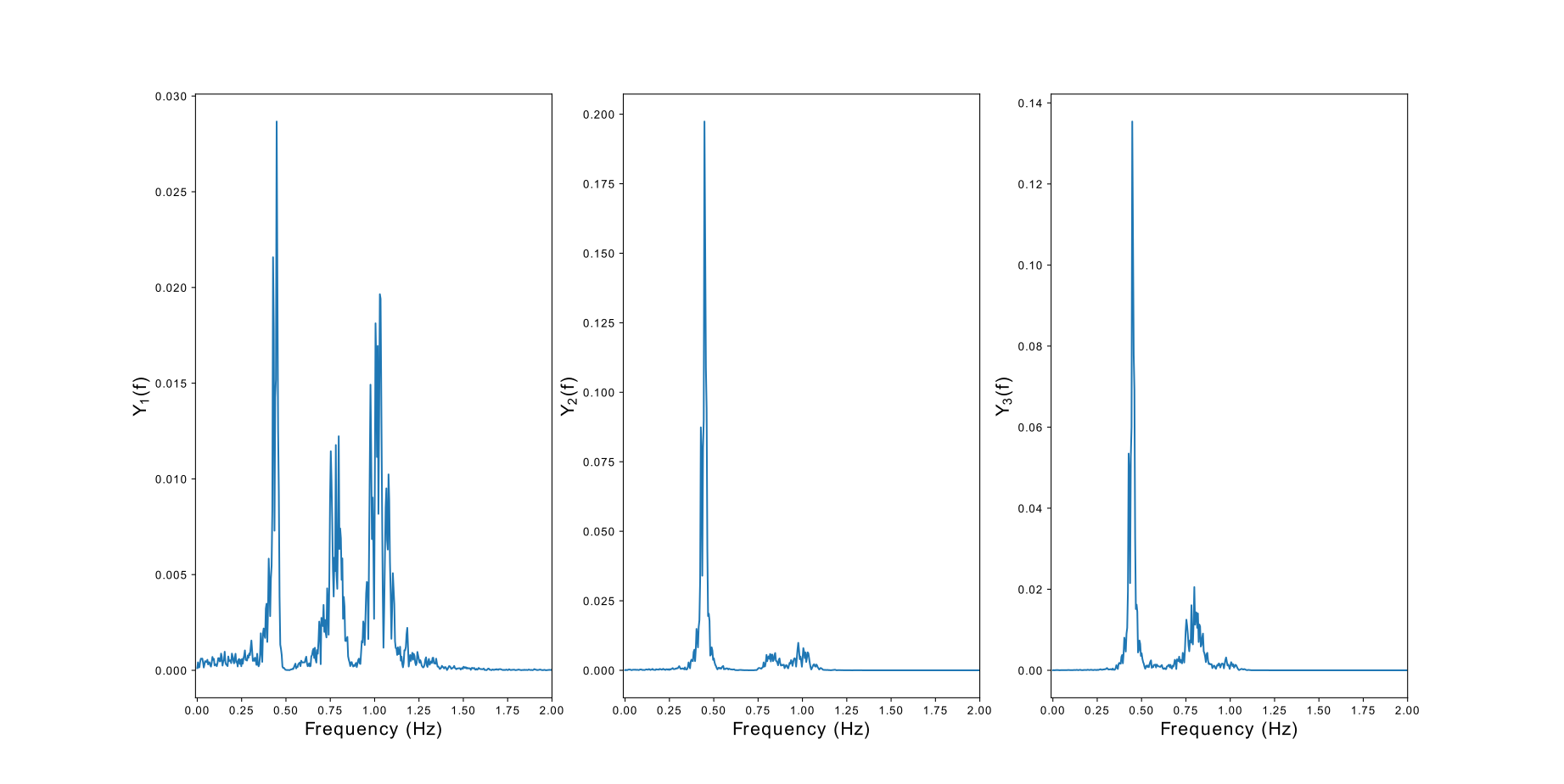}
    \end{adjustbox}
    \caption{PSDs of three-degree-of-freedom structure; physical coordinates.}
    \label{fig:physical_3DOF_PSD}
\end{figure}

\begin{figure}[H]
    \begin{adjustbox}{height=190pt, width=0.9\paperwidth, center}
    \includegraphics{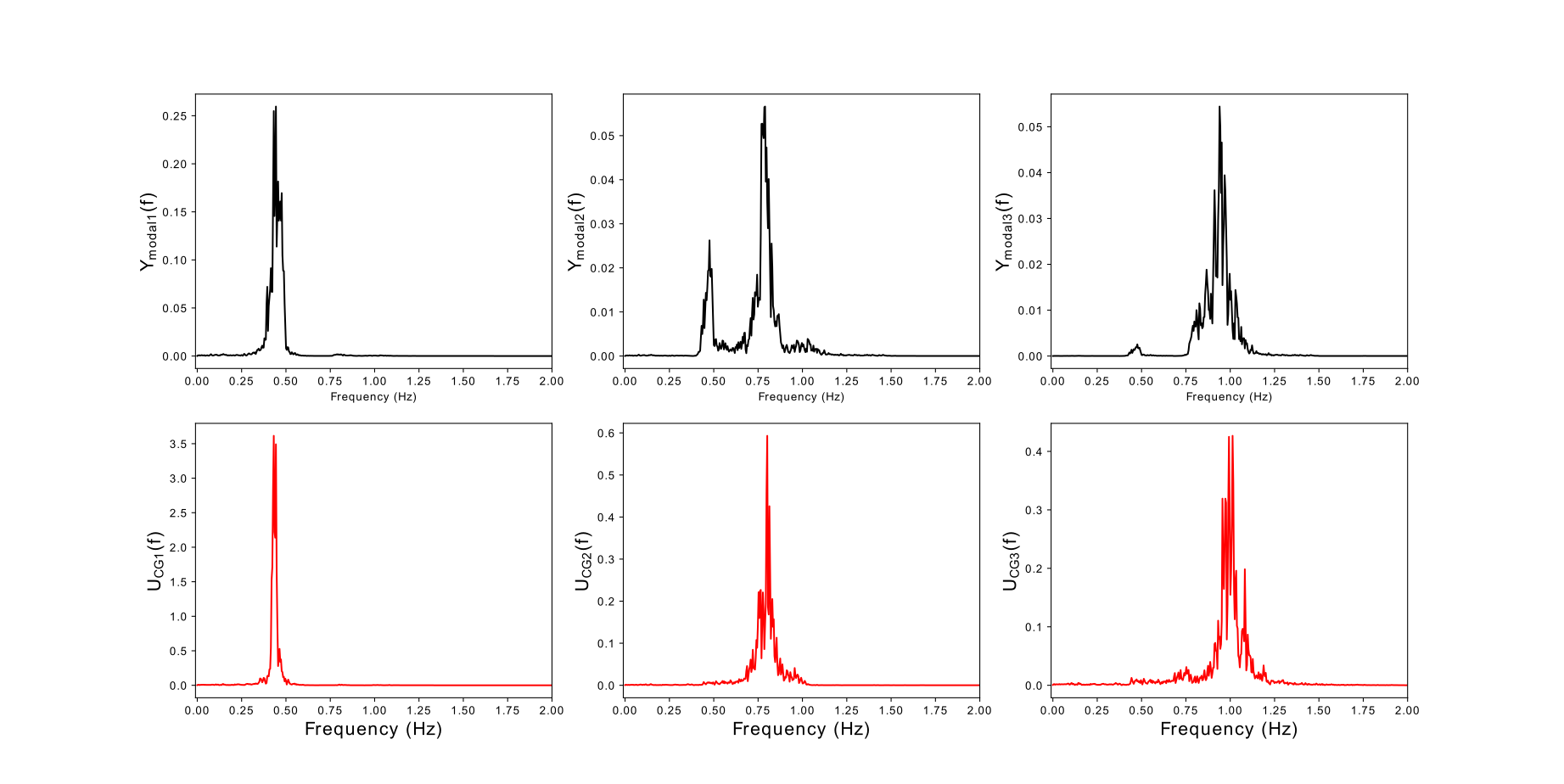}
    \end{adjustbox}
    \caption{PSDs of three-degree-of-freedom structure, linear modal decomposition (top) and cycle-GAN decomposition selected via the inner product criterion (bottom).}
    \label{fig:PCA_CG_3DOF_PSD}
\end{figure}

In order to test the method on a three-degree-of-freedom system, a third mass was added to the system shown in Figure \ref{fig:mass_spring}, connected to the second mass and the ground with a linear spring with $k=10$ and $c=0.1$. Exactly the same procedure was followed. The PSDs for the physical coordinates are shown in Figure \ref{fig:physical_3DOF_PSD}. The natural frequencies of the underlying linear system are $0.39$ Hz, $0.71$ Hz and $0.93$ Hz and again the spread and the movement because of the nonlinearity are seen. The best model for the current case study had $110$ hidden units and the decomposition provided by the algorithm in terms of PSDs is shown in the bottom row of Figure \ref{fig:PCA_CG_3DOF_PSD}. Again, for comparison, the linear modal/PCA coordinate PSDs are shown in the top row. The decoupling of the modes is clear. In every PSD a different peak is dominant and only small effects of other modes are present.

\subsection{Four-degree-of-freedom system}
\label{sec:4DOF}

\begin{figure}[H]
    \begin{adjustbox}{height=210pt, width=0.9\paperwidth, center}
    \includegraphics{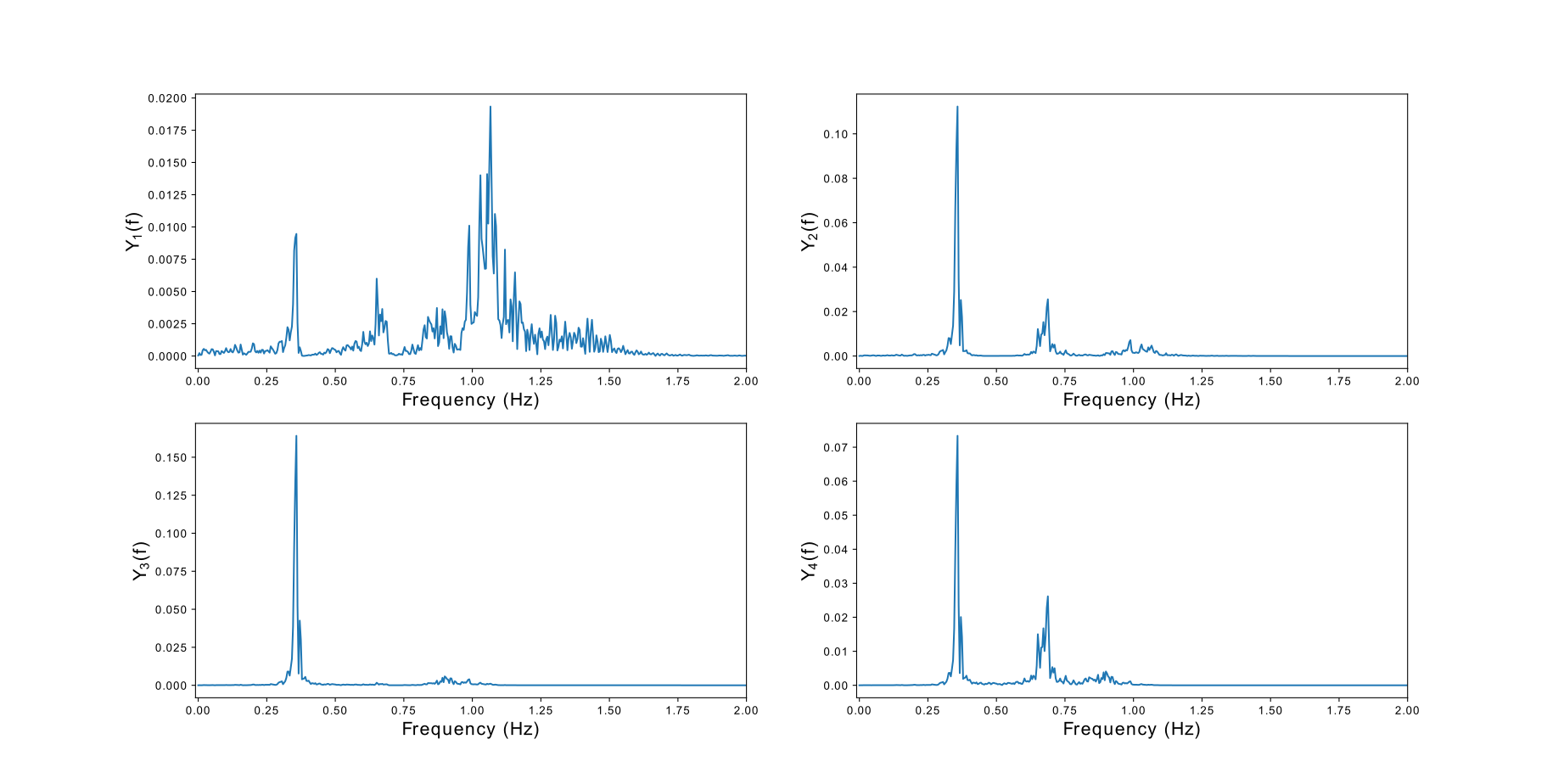}
    \end{adjustbox}
    \caption{PSDs of four-degree-of-freedom structure; physical coordinates.}
    \label{fig:physical_4DOF_PSD}
\end{figure}

\begin{figure}[H]
    \begin{adjustbox}{height=180pt, width=0.9\paperwidth, center}
    \includegraphics{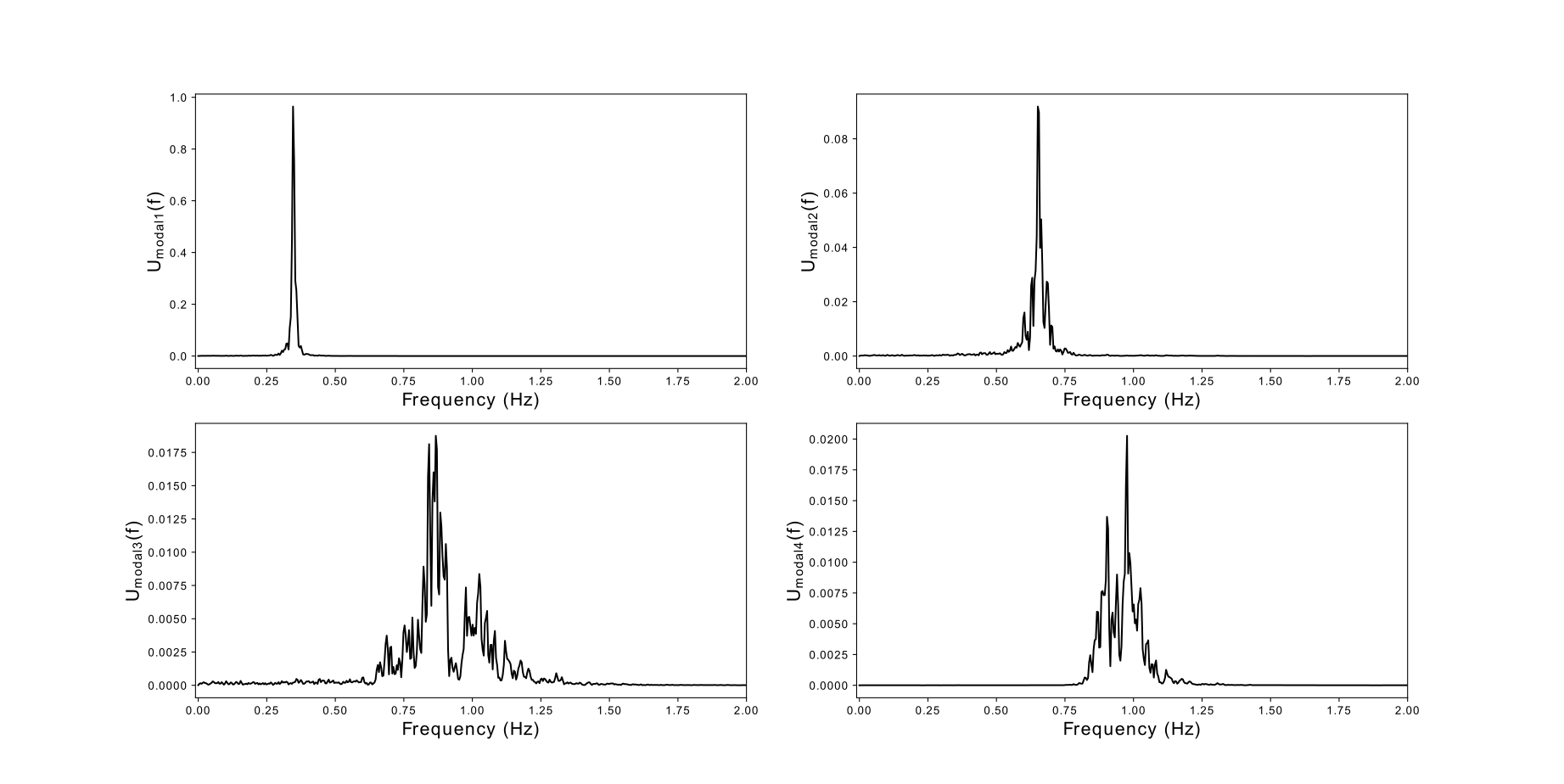}
    \end{adjustbox}
    \caption{PSDs of four-degree-of-freedom structure, linear modal decomposition coordinates.}
    \label{fig:PCA_4DOF_PSD}
\end{figure}

\begin{figure}[H]
    \begin{adjustbox}{height=180pt, width=0.9\paperwidth, center}
    \includegraphics{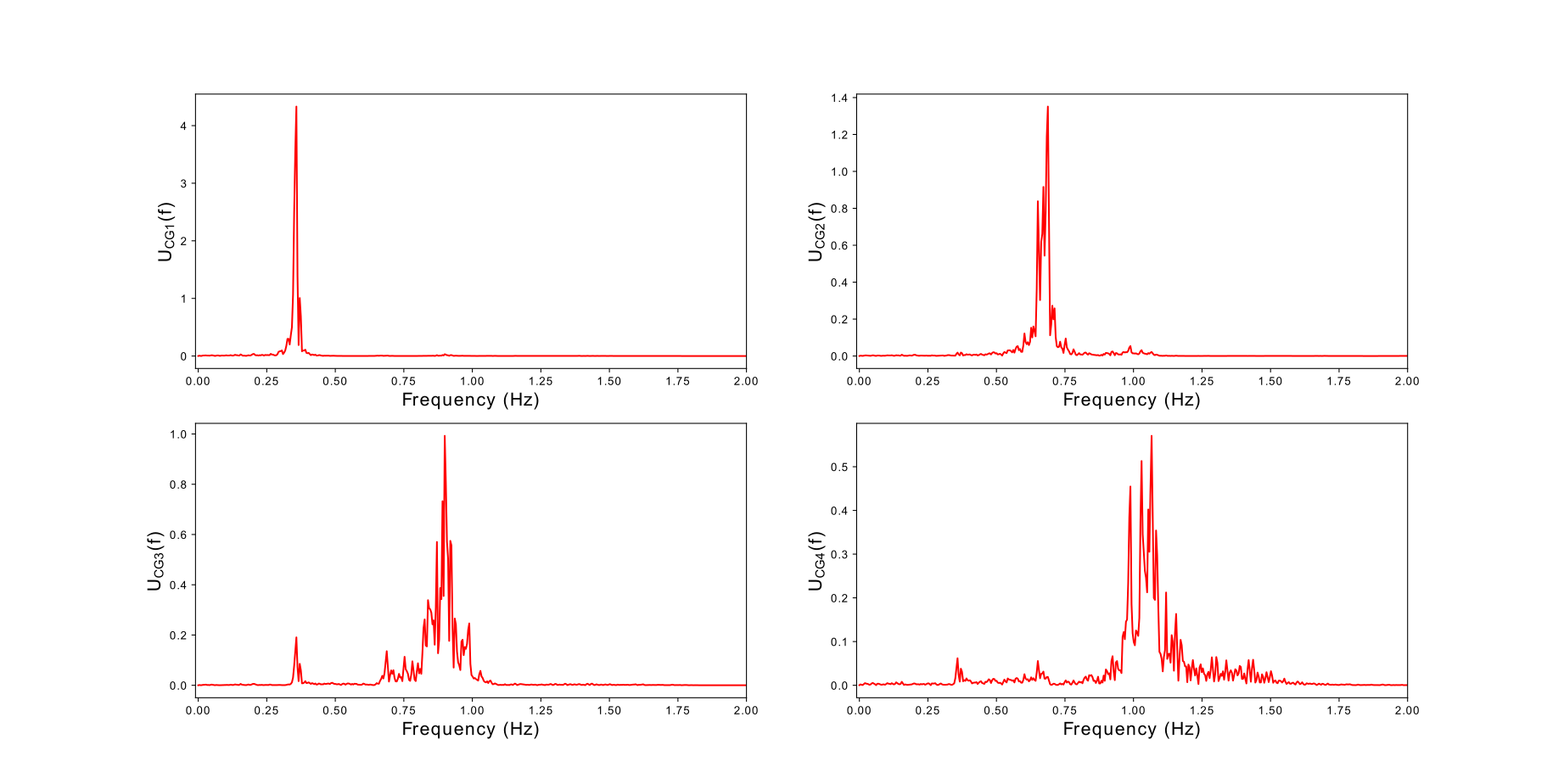}
    \end{adjustbox}
    \caption{PSDs of four-degree-of-freedom structure selected via the inner product criterion, cycle-GAN latent variables.}
    \label{fig:CG_4DOF_PSD}
\end{figure}

For the case of a four-degree-of-freedom system (which was not studied in \cite{worden2017machine}), it was noted that a larger value for the nonlinear stiffness parameter $k_3$ was needed in order for the structure to exhibit strongly-nonlinear behaviour; therefore, $k_3$ was increased to $3000$ for this case study. In this case the model that had the best performance was a neural network with $100$ neurons in its hidden layer. The PSDs of the four displacements are shown in Figure \ref{fig:physical_4DOF_PSD}. The natural frequencies of the underlying linear system are $0.31$ Hz, $0.59$ Hz, $0.81$ Hz and $0.96$ Hz but again a movement and a spreading towards higher frequencies is observed. In the first PSD, four peaks are clearly seen. It is expected from the decomposition to have single peaks for every modal coordinate PSD. Using a linear modal decomposition, single peaks are not achieved, as seen in Figure \ref{fig:PCA_4DOF_PSD} in the PSDs of the third and the fourth modal coordinates. However, using the proposed algorithm, in Figure \ref{fig:CG_4DOF_PSD}, a single peak can be seen as dominant in every modal coordinate PSD.

\subsection{Three-degree-of-freedom experimental system}
\label{sec:LANL}

\begin{center}
    \begin{figure}[H]
    \centering
    \includegraphics[scale=1.2]{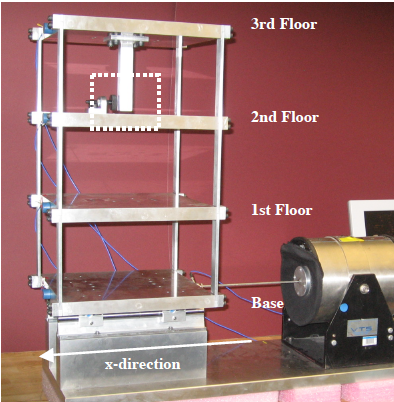}
    \caption{Experimental set-up of three-floor and the bumper nonlinearity between the second and third floor shown in the dashed box \cite{figueiredo2009structural}.}
    \label{fig:LANL_bookshelf}
\end{figure}
\end{center}

The experimental set-up for the data used in the current work was the one described in \cite{figueiredo2009structural}; the experimental structure is shown in Figure \ref{fig:LANL_bookshelf}. The structure was tested in $17$ different states some of them being considered damaged and some not. The different states and their description are shown in Table \ref{Tab:structural_states}. Particularly interesting are the states $10$-$14$. In these states, a column between the second and the third floor is placed near a bumper, introducing a nonlinearity into the system in the form of bilinear stiffness. Each state has a different value for the gap between the column and the bumper (from $0.20$mm to $0.5$mm). For the analysis here, the data from State $12$ were used, where the gap was $0.13$ mm, as well as from State $14$, where the gap was $0.05$ mm. The latter case was severely nonlinear because of the more frequent collisions. These cases may be considered harshly nonlinear compared to the smooth nonlinearities of the simulated case studies. The structure had four sensors recording accelerations, one on the base and one on each floor. For the current work, the displacements of the three floors are used, considering the base acceleration as the excitation (as in an earthquake).

\newpage

\vfill
\vfill
\newpage

\begin{table}[H]
    \centering
    \begin{tabular}{ |p{1.9cm}|p{2.1cm}|p{7.0cm}|}
        \hline
        \thead{Label} & \thead{State \\ Condition} & \thead{Description}\\
        \hline
        State \#1   & Undamaged & Baseline condition\\
        State \#2   & Undamaged & Added mass (1.2 kg) at the base\\
        State \#3   & Undamaged & Added mass (1.2 kg) on the $1^{st}$ floor\\
        State \#4   & Undamaged & Stiffness reduction in base front column\\
        State \#5   & Undamaged & Stiffness reduction in base front and rear column\\
        State \#6   & Undamaged & Stiffness reduction in $1^{st}$ floor front column\\
        State \#7   & Undamaged & Stiffness reduction in $1^{st}$ floor front and rear column\\
        State \#8   & Undamaged &  Stiffness reduction in $2^{nd}$ floor front column\\
        State \#9   & Undamaged &  Stiffness reduction in $2^{nd}$ floor front and rear column\\
        State \#10   & Damaged &  Gap ($0.20$ mm)\\
        State \#11   & Damaged &  Gap ($0.15$ mm)\\
        State \#12   & Damaged &  Gap ($0.13$ mm)\\
        State \#13   & Damaged &  Gap ($0.10$ mm)\\
        State \#14   & Damaged &  Gap ($0.05$ mm)\\
        State \#15   & Damaged &  Gap ($0.20$ mm) and mass ($1.2$ kg) at the base\\
        State \#16   & Damaged &  Gap ($0.20$ mm) and mass ($1.2$ kg) at the $1^{st}$ floor\\
        State \#17   & Damaged &  Gap ($0.10$ mm) and mass ($1.2$ kg) at the $1^{st}$ floor\\
        \hline
    \end{tabular}
    \caption{Description of different states of the experimental set-up \cite{figueiredo2009structural}.}
    \label{Tab:structural_states}
\end{table}	

\subsubsection{Experimental data: State 12}

\begin{figure}[H]
    \begin{adjustbox}{height=180pt, width=0.8\paperwidth, center}
    \includegraphics{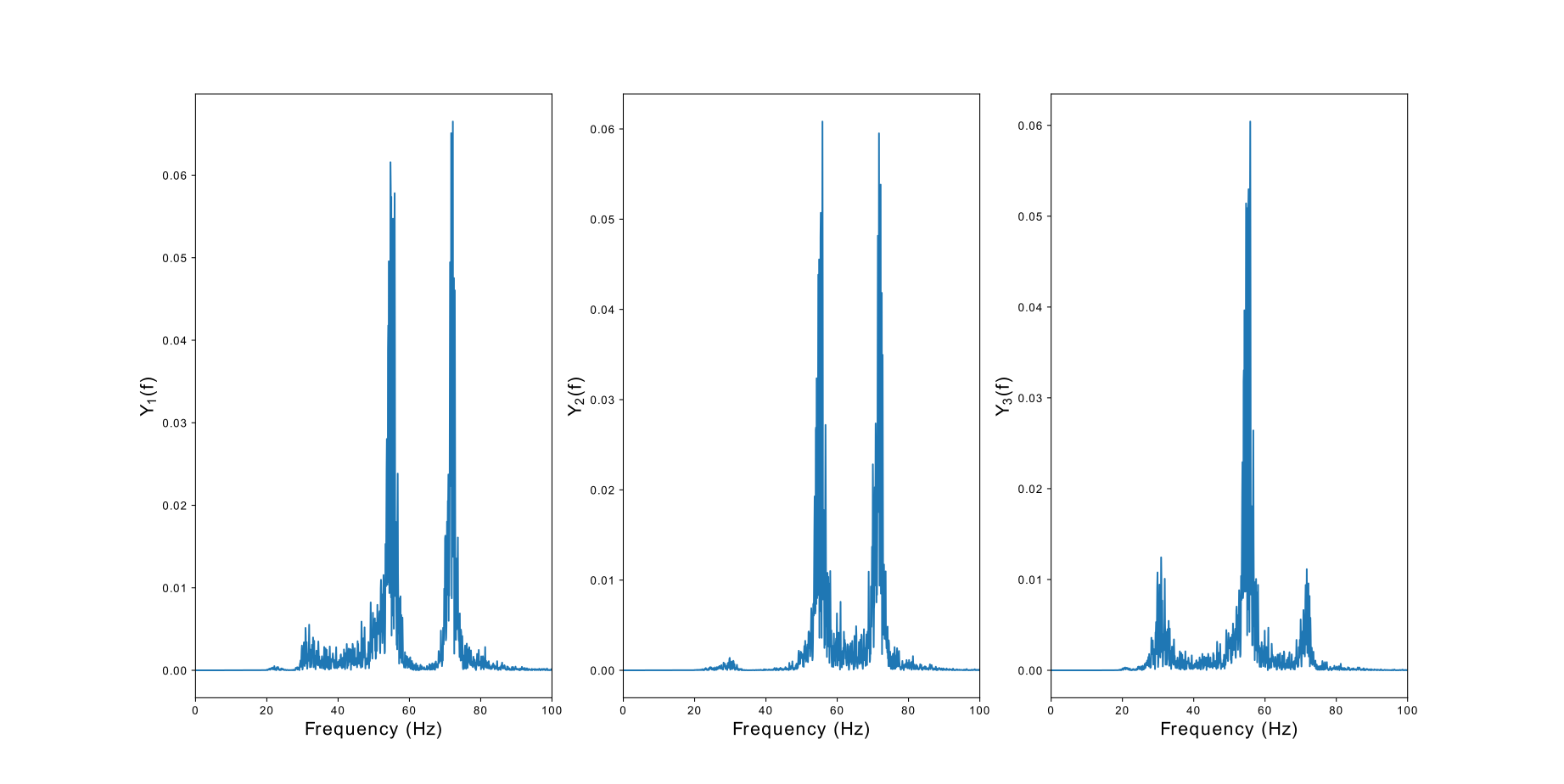}
    \end{adjustbox}
    \caption{PSDs samples of three-floor experimental structure, physical coordinates of state 12.}
    \label{fig:physical_LANL_PSD}
\end{figure}

Similar to the simulated examples, the PSDs of the signals of the three sensors are shown in Figure \ref{fig:physical_LANL_PSD}. The PSDs in the specific figure are computed from one out of $50$ experiments performed corresponding to State $12$. Three peaks can clearly be seen. For the rest of the experimental case studies presented, the PSDs are calculated as the average PSD of the $50$ experiments performed.

\begin{figure}[H]
    \begin{adjustbox}{width=0.9\paperwidth, center}
    \includegraphics{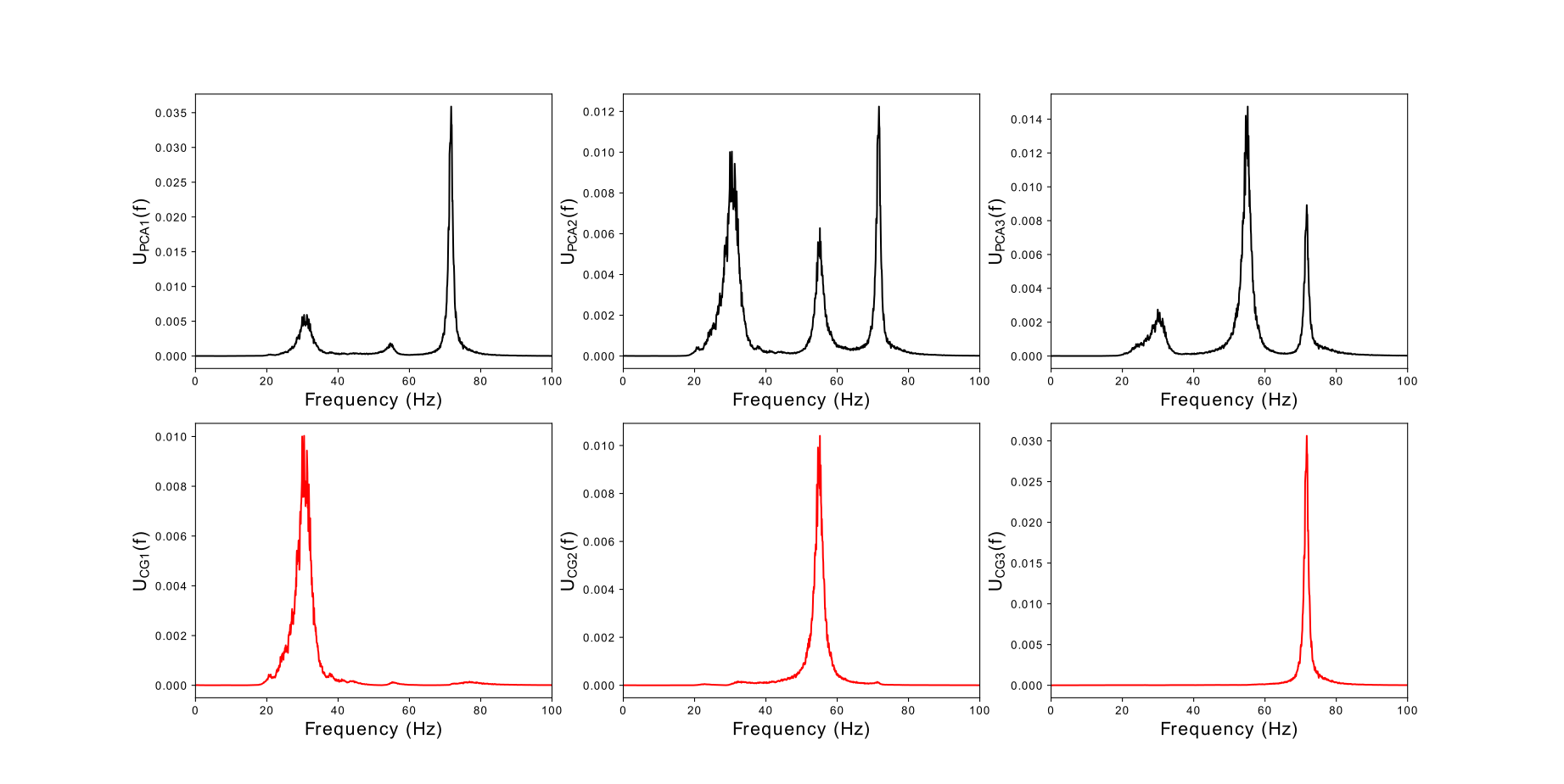}
    \end{adjustbox}
    \caption{PSDs of three-floor experimental structure, PCA decomposition (top) and cycle GAN decomposition selected via the inner product criterion (bottom); state 12.}
    \label{fig:PCA_CG_LANL_PSD}
\end{figure}

Training the cycle-GAN using the accelerations from all $50$ experiments, the decomposition achieved is shown in the bottom plots of Figure \ref{fig:PCA_CG_LANL_PSD}. The model that yielded the results shown had $100$ nodes in its hidden layers. The decoupling of the modes is clear. Each latent variable corresponds to a different mode and the algorithm performs better than using a PCA decomposition (shown in the top row of the same figure). Each latent variable's PSD has a clearly dominant peak. In the first, and especially in the second, plot of the PCA decomposition PSDs, the modes are coupled and none is dominant.

\subsubsection{Experimental data: State 14}

\begin{figure}[H]
    \begin{adjustbox}{height=190pt, width=0.9\paperwidth, center}
    \includegraphics{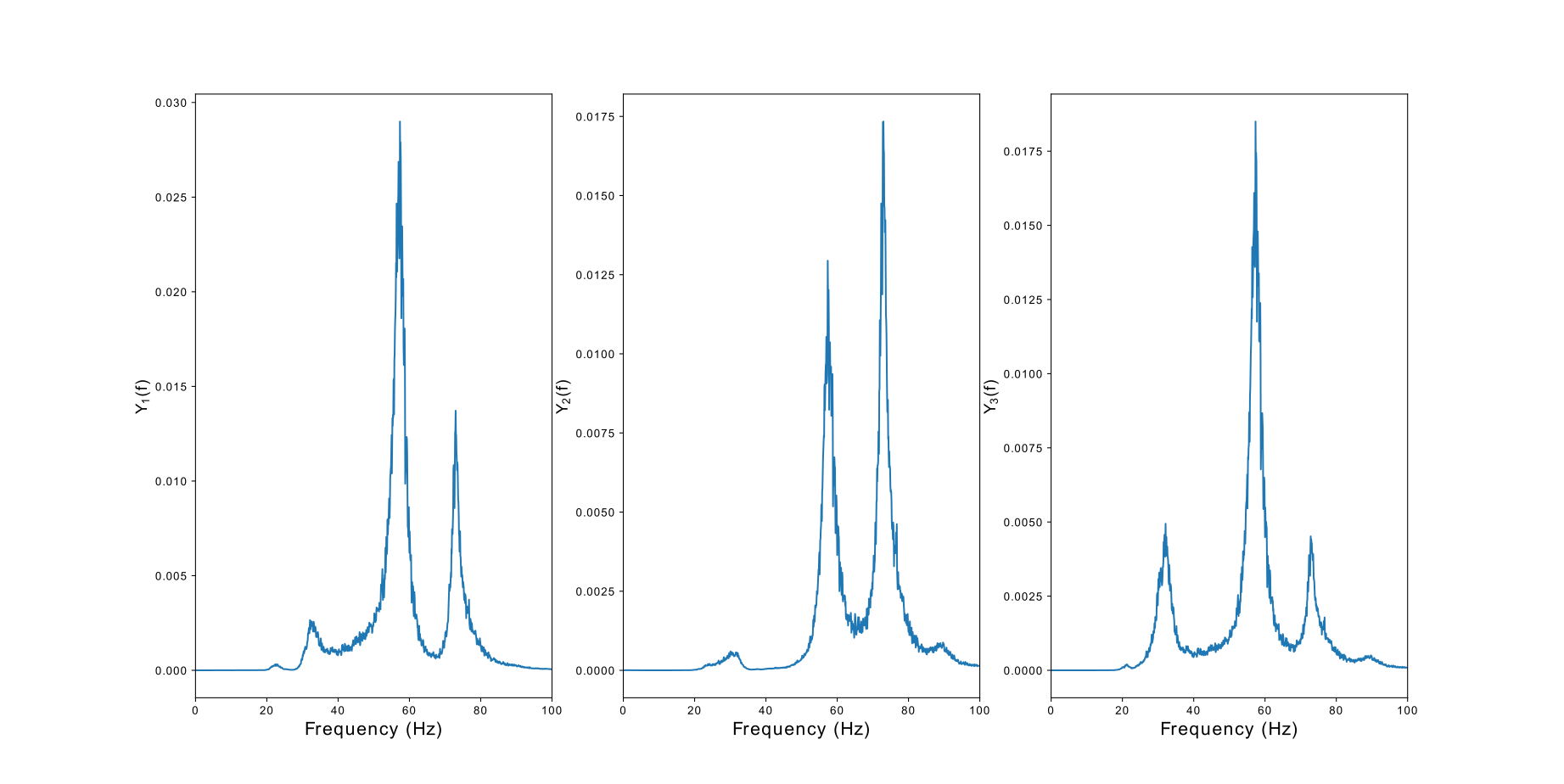}
    \end{adjustbox}
    \caption{Average PSDs of three-floor experimental structure, physical coordinates of state 14.}
    \label{fig:physical_LANL_PSD_state_14}
\end{figure}

\begin{figure}[H]
    \begin{adjustbox}{width=0.9\paperwidth, center}
    \includegraphics{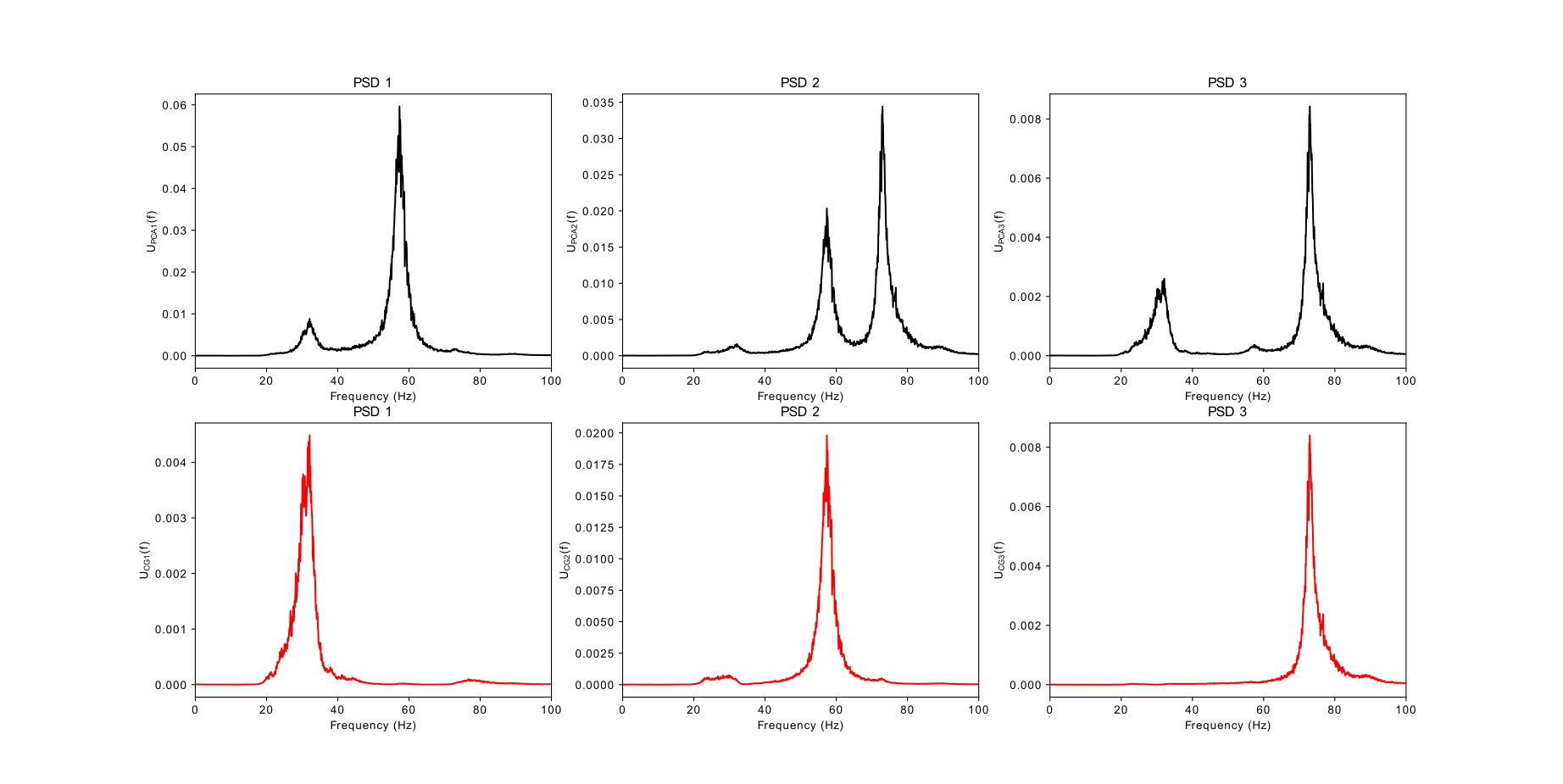}
    \end{adjustbox}
    \caption{PSDs of three-floor experimental structure, PCA decomposition (top) and cycle GAN decomposition selected via the inner product criterion (bottom); state 14.}
    \label{fig:PCA_CG_LANL_PSD_state_14}
\end{figure}

Following the same procedure, using the State $14$ data (whose natural coordinates PSDs are shown in Figure \ref{fig:physical_LANL_PSD_state_14}), the results of the modal decomposition are shown in Figure \ref{fig:PCA_CG_LANL_PSD_state_14}. The comparison between the cycle-GAN decomposition and the PCA decomposition clearly indicate that the former has achieved better results. The separation of the modes is almost perfect compared to the PCA case, where the yielded coordinates clearly do not achieve any modal separation.

\section{Superposition}
\label{sec:superposition}

The second essential part of modal analysis is the superposition step. For linear modal analysis, superposition is defined simply as the summation of the displacements that correspond to each mode. In the case of nonlinear modes, clearly it cannot be that simple. A nonlinear superposition function has to be defined in order to map coordinates from the modal space back to the physical. Definition of a superposition function, in the current work, is a major addition to the method compared to \cite{worden2017machine, dervilis2019nonlinear}.

Since the cycle-GAN was used, this mapping has already been defined. The inverse mapping is the goal of the second generator. The second generator has been trained in parallel with the generator that was used to decompose the movement of the structure into the modal space. Subsequently, the performance of the inverse mapping for the presented case studies is shown and discussed.

For each case study, the superposition mapping is provided by the generator $G_{U \to Y}$. The generator selected is taken from the same training epoch as the forward mapping generator $G_{Y \to U}$. To evaluate the superposition efficiency, a \textit{normalised mean-square error} (NMSE) for each reconstruction is computed by,
\begin{equation}
    \centering
    \label{eq:NMSE}
    NMSE = \frac{100}{N\sigma_{y}^{2}}\sum_{i=1}^{N}(\hat{y}_{i} - y_{i})^{2}
\end{equation}
where $N$ is the total number of displacements samples, $\sigma_{y}^{2}$ is the variance of the displacements, $y_{i}$ is the real recorded displacement and $\hat{y}_{i}$ is the superposition provided by $G_{Y \to U}(G_{U \to Y}(y))$. The NMSE in each case study is computed using all the displacement samples from every degree of freedom of the systems. The NMSE is a convenient measure of error in regression problems, since it provides an objective measure of the accuracy, regardless of the scale of the data. NMSE values close to $100\%$ indicate that the model does no better than simply using the mean value of the data, while the lower the value the better the model is calibrated. From experience, values of NMSE lower than $5\%$ indicate a well-fitted model, and values lower than $1\%$ show an excellent model.

\subsection{Superposition for the two-degree-of-freedom system}
\label{sec:2DOF_super}

\begin{figure}[H]
    \begin{adjustbox}{height=190pt, width=0.68\paperwidth, center}
    \includegraphics{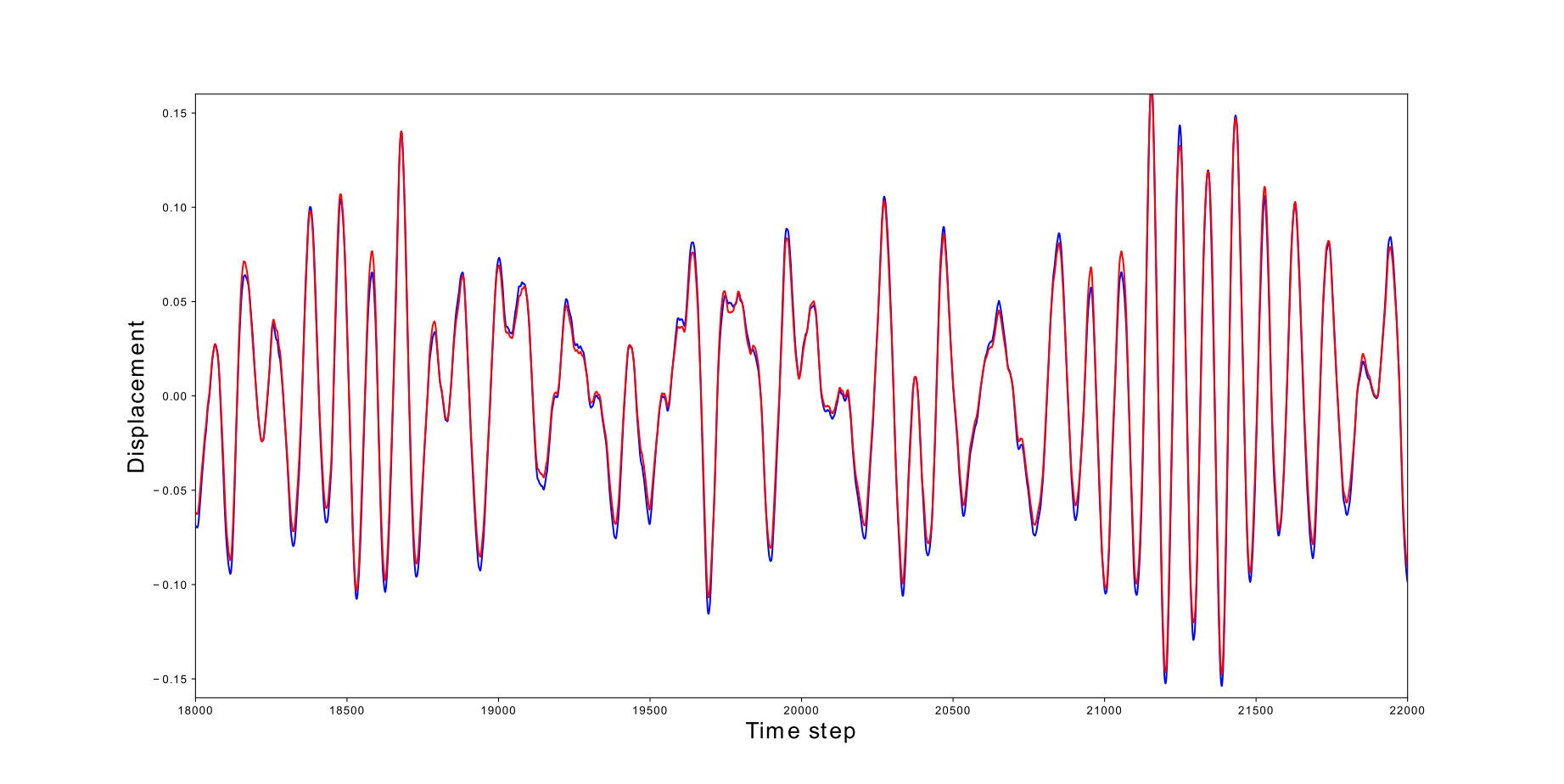}
    \end{adjustbox}
    \caption{Superposition/inverse modal transformation for the two-degree-of-freedom system (red) and original displacements (blue).}
    \label{fig:super_2DOF}
\end{figure}

Part of the results of superposition for the two-degree-of-freedom system are shown in Figure \ref{fig:super_2DOF}. The inverse-mapping function provided by the generator $G_{Y \to U}$ is very accurate; This is confirmed by the NMSE value, which for this case is $0.46\%$.

\subsection{Superposition for the three-degree-of-freedom system}
\label{sec:3DOF_super}

\begin{figure}[H]
    \begin{adjustbox}{height=190pt, width=0.68\paperwidth, center}
    \includegraphics{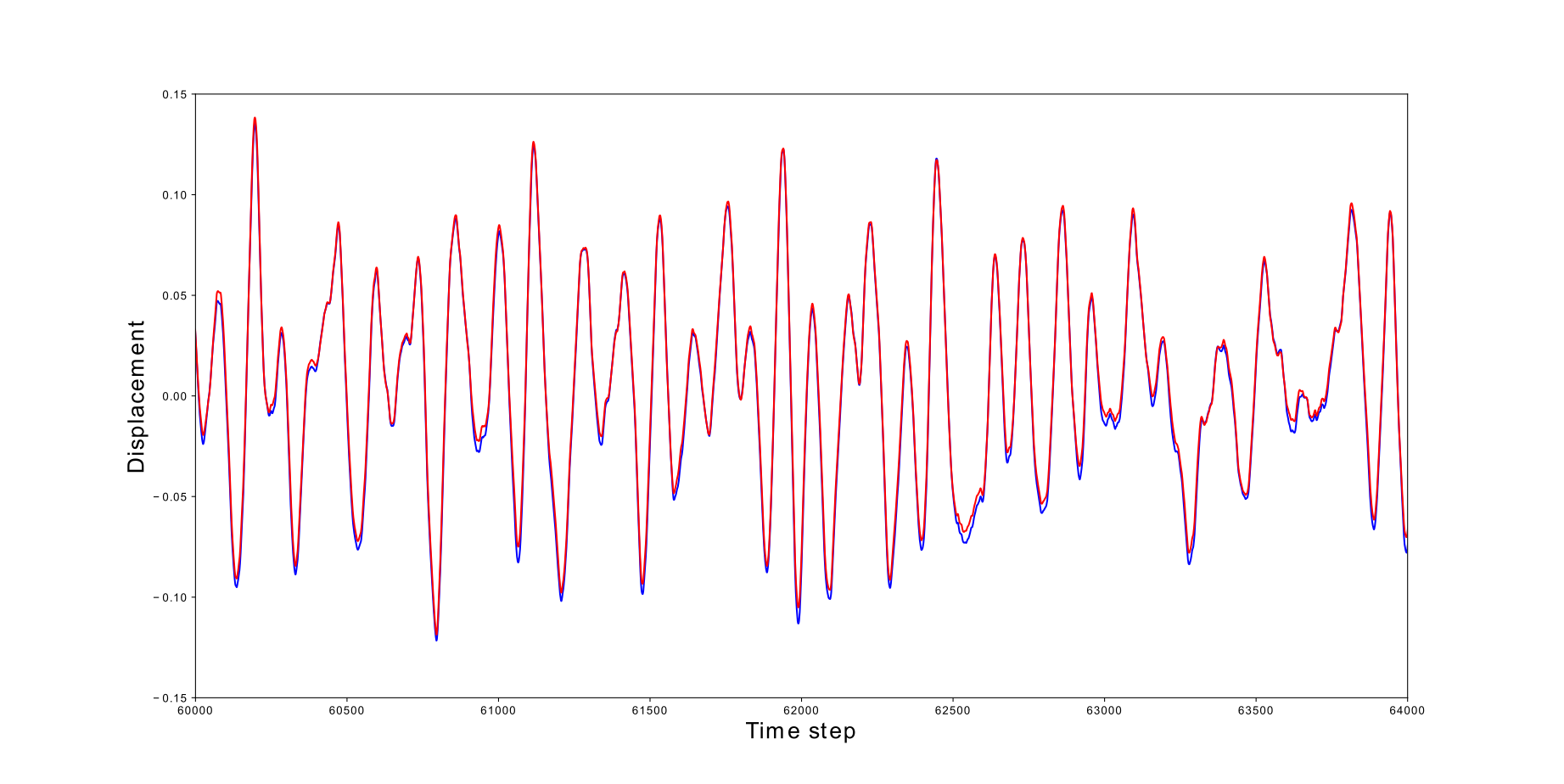}
    \end{adjustbox}
    \caption{Superposition/inverse modal transformation for the three-degree-of-freedom system(red) and original displacements (blue).}
    \label{fig:super_3DOF}
\end{figure}

For the three-degree-of-freedom system, some results are shown in Figure \ref{fig:super_3DOF}. The inverse mapping this time seems a little less accurate visually; however, the NMSE value for this case is still only $0.294\%$.

\subsection{Superposition for the four-degree-of-freedom system}
\label{sec:4DOF_super}

\begin{figure}[H]
    \begin{adjustbox}{height=190pt, width=0.68\paperwidth, center}
    \includegraphics{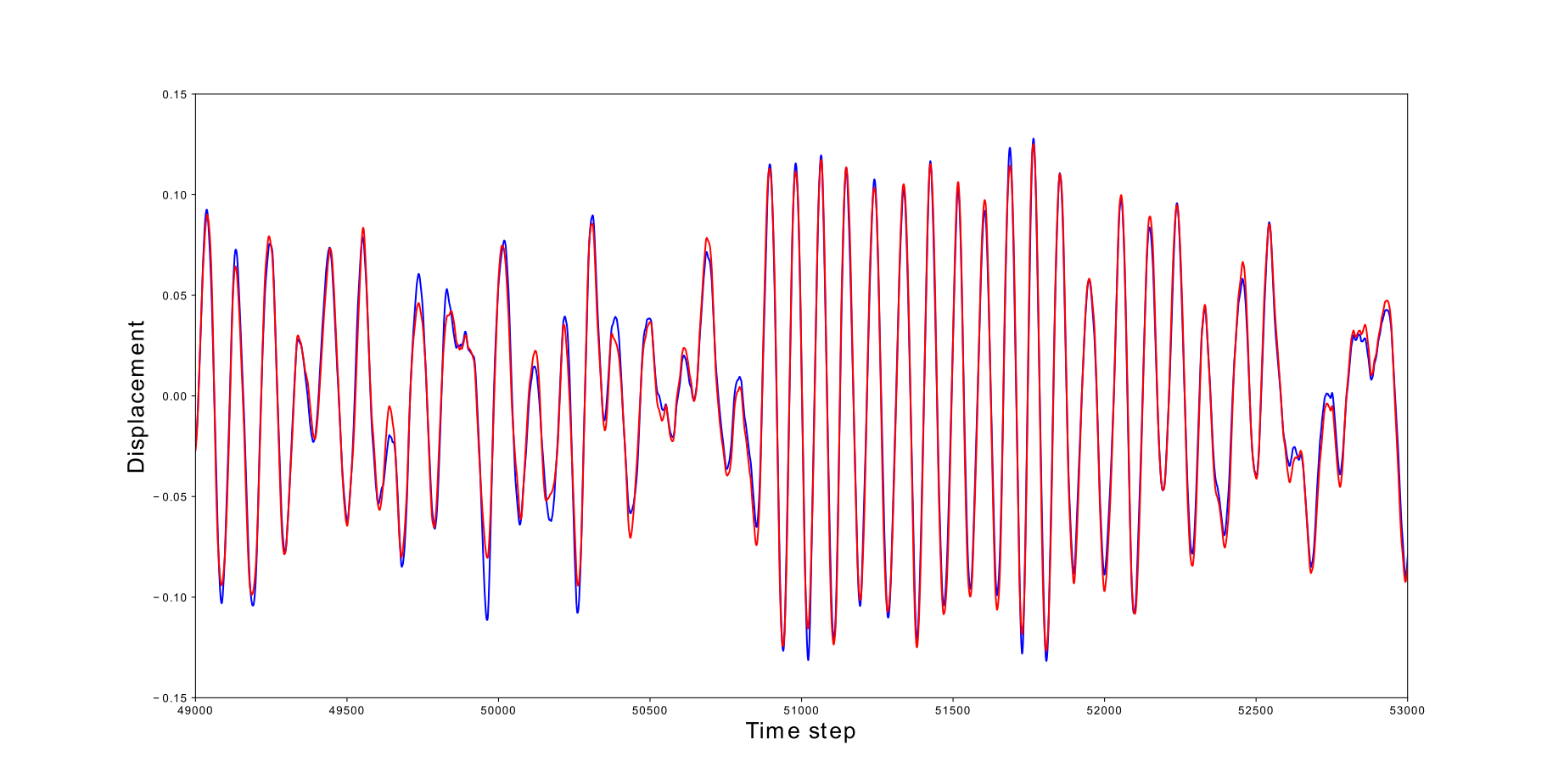}
    \end{adjustbox}
    \caption{Superposition/inverse modal transformation for the four-degree-of-freedom system(red) and original displacements (blue).}
    \label{fig:super_4DOF}
\end{figure}

For the four-degree-of-freedom system, some results are shown in Figure \ref{fig:super_4DOF}. The NMSE value for this case is $1.92\%$. This error is higher than the previous cases, but still implies a good inverse mapping.

\subsection{Superposition for the experimental system}
\label{sec:LANL_super}

\subsubsection{Experimental data: State 12}

\begin{figure}[H]
    \begin{adjustbox}{height=190pt, width=0.68\paperwidth, center}
    \includegraphics{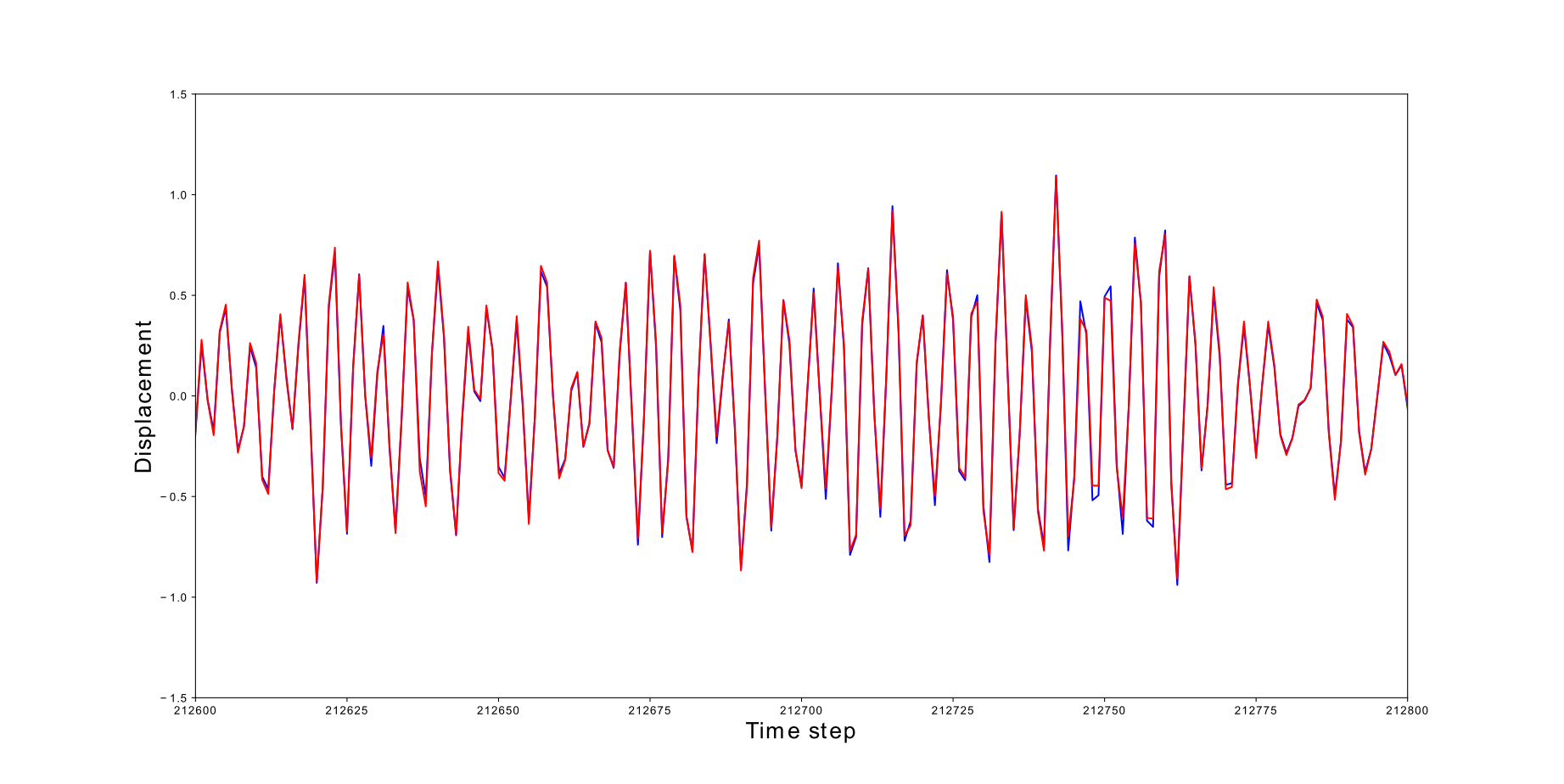}
    \end{adjustbox}
    \caption{Superposition/inverse modal transformation for the experimental system(red) and original displacements (blue), state 12.}
    \label{fig:super_LANL}
\end{figure}

For the State 12 of the experimental system, some results are shown in Figure \ref{fig:super_LANL}; the NMSE value for this case is $1.23\%$.

\subsubsection{Experimental data: State 14}

\begin{figure}[H]
    \begin{adjustbox}{height=190pt, width=0.68\paperwidth, center}
    \includegraphics{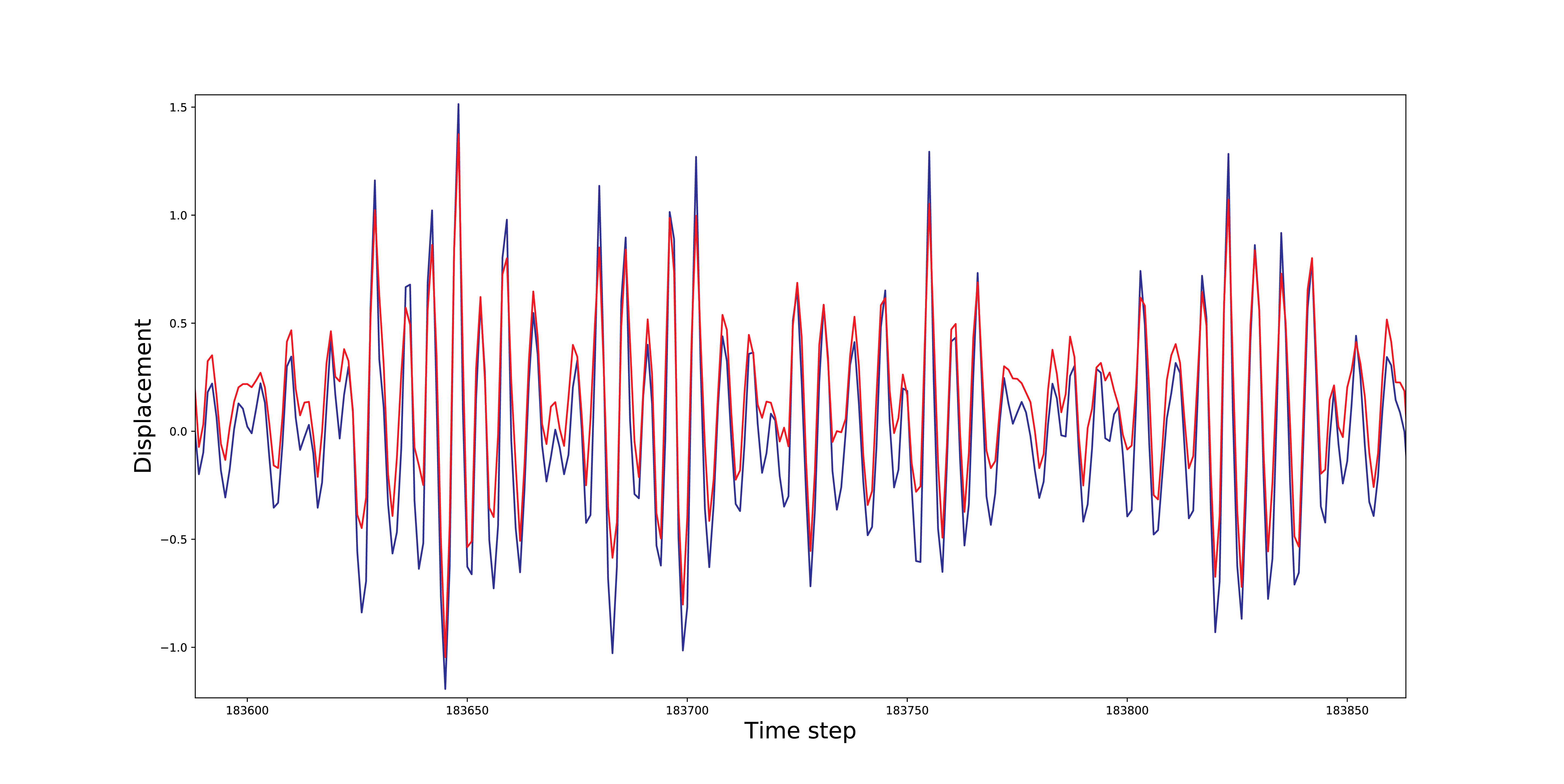}
    \end{adjustbox}
    \caption{Superposition/inverse modal transformation for the experimental system(red) and original displacements (blue), state 14.}
    \label{fig:super_LANL_state_14}
\end{figure}

For the State 14 of the experimental system, some results are shown in Figure \ref{fig:super_LANL_state_14}. The NMSE value for this case is $10.93\%$. Clearly the performance of the algorithm is not as good as in the previous examples. This might be explained by the harsh and highly-nonlinear nature of this final state of the experimental set up.

\section{Modal correlation study}
\label{sec:cor_study}

In a previous section it was mentioned that modal coordinates are expected to be independent, since the modal space is pre-defined as sampling from variables $Y \sim \mathcal{N}(\bm{\mu}, I_{n})$. However, the algorithm is searching for a mapping from the modal space to the natural without any constraint regarding which samples of the modal space to use. At the same time, the algorithm has to balance three types of losses (adversarial, reconstruction and orthogonality), and may result in overlooking one of them in some degree. The adversarial loss, as discussed, enforces the statistical independence of the modal coordinates and, if overlooked in some degree, the result may be that the modal coordinates are correlated. In the current section, a further study of the correlation of the modal coordinates is performed.

\subsection{Linear correlation}
\label{sec:linear_cor}

The most simple case of correlation measure here will be Pearson's correlation coefficient, which for a pair of random variables $(X, Y)$ is given by,

\begin{equation}
    \rho_{X, Y} = \frac{\mathbb{E}[(X - \mu_{X})(Y - \mu_{Y})]}{\sigma_{X}\sigma_{Y}}
\end{equation}
where $\mu_{X}$, $\mu_{Y}$ are the mean values of $X$ and $Y$ respectively, $\mathbb{E}[\ ]$ is the expected value of the quantity in brackets and $\sigma_{X}$, $\sigma_{Y}$ the standard deviations of $X$ and $Y$. The correlation coefficient $\rho$ takes values in the interval $[-1, 1]$. Values closer to $-1$ and $1$ mean that the samples are highly correlated, while values closer to $0$ mean that the two random variables are less correlated (in a linear manner).

Using such a correlation measure between observations of two variables reveals only potential linear dependency between the variables. For linear modal analysis, such a measure is minimised for the modal coordinates. Since linear modal analysis is equivalent to PCA, this is an expected effect of the transformation. Therefore, to study correlation in the case of nonlinear modal analysis, a nonlinear correlation measure should also be used.

\subsection{Nonlinear correlation}
\label{sec:nonlinear_cor}

The correlation measure selected here to study the nonlinear modes is \textit{distance correlation} \cite{szekely2013distance}. The distance correlation can be thought of as a generalisation of Pearson's correlation coefficient, and is a way of detecting higher-order correlations between data. In order to calculate the distance correlation for two random variables $(X, Y)$, two distance matrices $A$ and $B$ have to be defined. To do this, elements $\alpha_{j,k}$ and $\beta_{j,k}$ are first defined by,
\begin{equation}
    \begin{split}
    \alpha_{j,k} = \norm{x_{j} - x_{k}}, \qquad k,j= 1, 2, ..., n \\
    \beta_{j,k} = \norm{y_{j} - y_{k}}, \qquad k,j= 1, 2, ..., n
    \end{split}
\end{equation}
where $n$ is the number of observations. Subsequently, the matrices $A$ and $B$ are computed as,
\begin{equation}
    \begin{split}
    A_{j,k} = \alpha_{j,k} - \Bar{\alpha}_{j\cdot} - \Bar{\alpha}_{\cdot k} + \Bar{\alpha}_{\cdot \cdot}, \qquad k,j= 1, 2, ..., n \\
    B_{j,k} = \beta_{j,k} - \Bar{\beta}_{j\cdot} - \Bar{\beta}_{\cdot k} + \Bar{\beta}_{\cdot \cdot}, \qquad k,j= 1, 2, ..., n
    \end{split}
\end{equation}
where $\Bar{\alpha}_{\cdot k}$ is the mean value of the $k$th column, $\Bar{\alpha}_{j \cdot}$ the mean of the $j$th row and $\Bar{\alpha}_{\cdot \cdot}$ the mean value of all $\alpha$ elements. Having computed the two matrices, the \textit{distance covariance} is given by,
\begin{equation}
    dCov^{2}(X, Y) = \frac{1}{n^{2}}\sum_{j=1}^{n}\sum_{k=1}^{n} A_{j,k} B_{j,k}
\end{equation}
Defining as \textit{distance variance} the distance covariance of a variable to itself ($dVar^{2}(X) = dCov^{2}(X, X)$) the distance correlation is calculated by,
\begin{equation}
    dCor(X, Y) = \frac{dCov(X,Y)}{\sqrt{dVar(X) dVar(Y)}}
\end{equation}

The distance correlation measure is bounded in the interval $[0, 1]$. Lower values indicate independence of the random variables while higher values indicate higher dependence. In the following subsections, both correlation metrics will be computed for the modal coordinates defined by the algorithm proposed here. 

\subsection{Correlation in the case studies}
\label{sec:corr}

\subsubsection{Correlation of the two-degree-of-freedom modes}
\begin{figure}[H]
    \centering
    \begin{adjustbox}{width=0.7\paperwidth, center}
    \begin{subfigure}[b]{0.49\textwidth}
        \includegraphics[width=.90\textwidth]{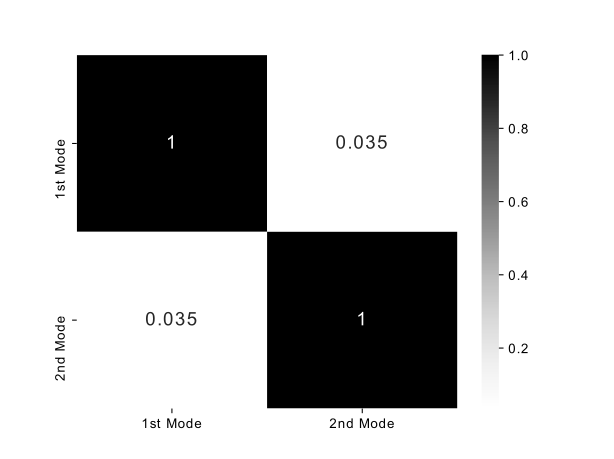}
    \end{subfigure}
    \begin{subfigure}[b]{0.49\textwidth}
        \includegraphics[width=.90\textwidth]{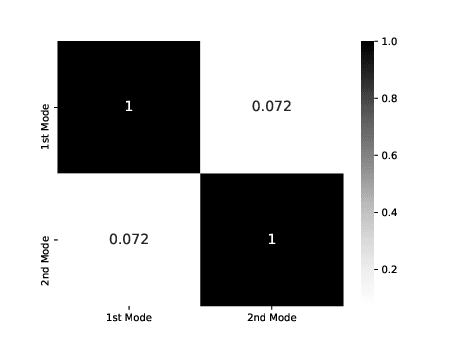}
    \end{subfigure}
    \end{adjustbox}
    \caption{Pearson's linear correlation coefficient (left) and distance correlation coefficient (right) of the modal decomposition computed for the two-degree-of-freedom system.}
    \label{fig:2d_corrs}
\end{figure}

Calculating the correlation coefficients for the modal coordinates of the two-degree-of-freedom system, the results are plotted in a `heat-map' coloured matrix in Figure \ref{fig:2d_corrs}. On the diagonal, obviously, the correlation coefficients are equal to $1$ since they are the autocorrelation coefficients of the modal coordinates. The values of the coefficients between the two modes are quite low indicating that independence of the modes has been achieved to a satisfactory degree.

\subsubsection{Correlation of three-degree-of-freedom modes}
\label{sec:3D_corr}

\begin{figure}[H]
    \centering
    \begin{adjustbox}{width=0.7\paperwidth, center}
    \begin{subfigure}[b]{0.49\textwidth}
        \includegraphics[width=.90\textwidth]{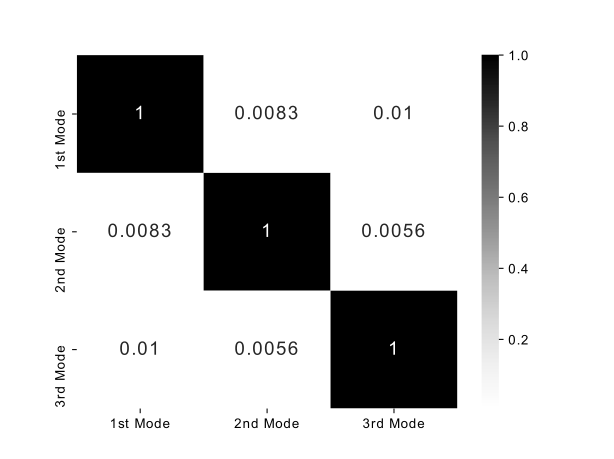}
    \end{subfigure}
    \begin{subfigure}[b]{0.49\textwidth}
        \includegraphics[width=.90\textwidth]{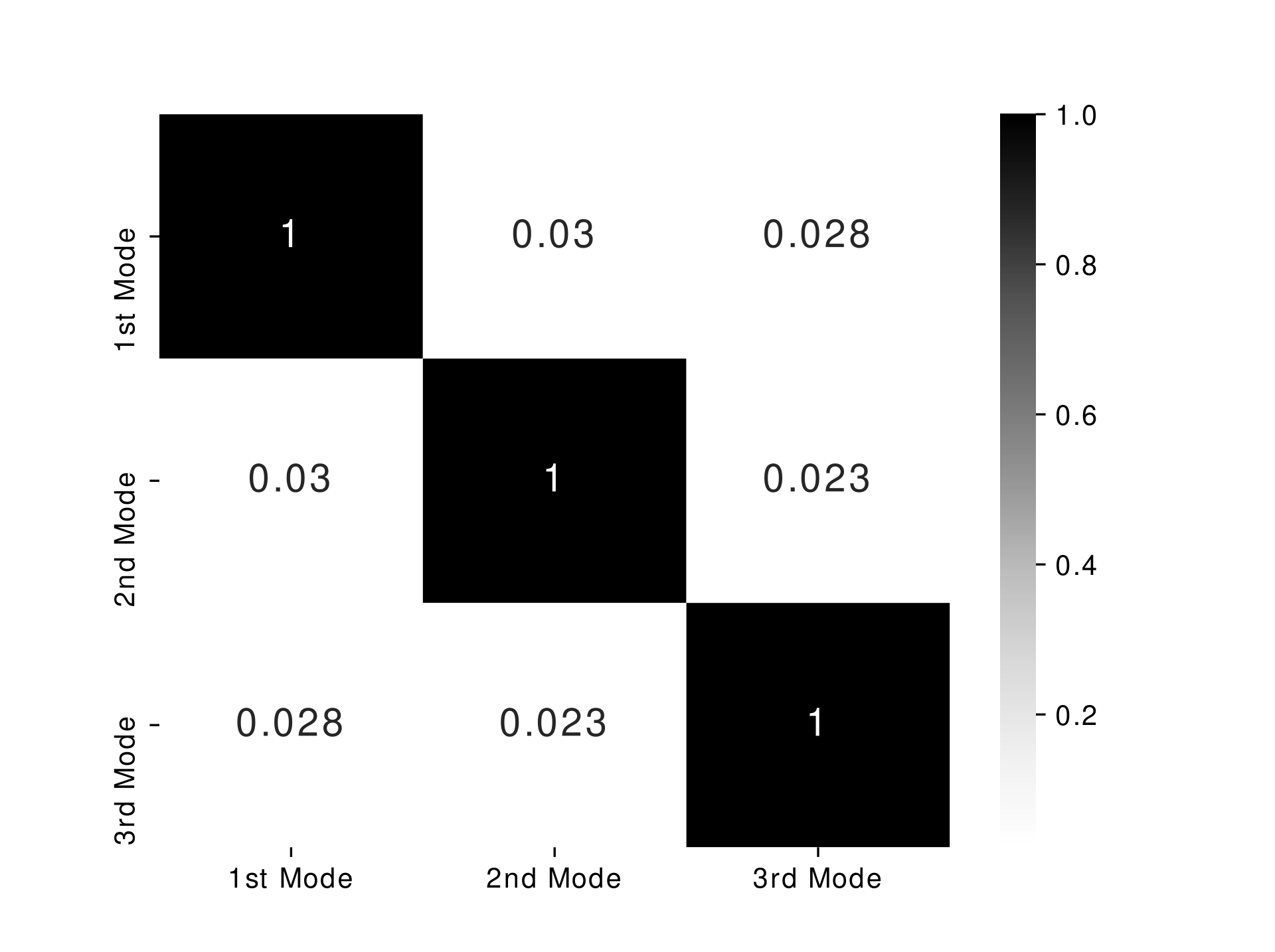}
    \end{subfigure}
    \end{adjustbox}
    \caption{Pearson's linear correlation coefficient (left) and distance correlation coefficient (right) of the modal decomposition computed for the three-degree-of-freedom system.}
    \label{fig:3d_corrs}
\end{figure}

For the three-degree-of-freedom system, the correlation coefficients are shown in Figure \ref{fig:3d_corrs}. In this case the correlation coefficients are also quite low.

\subsubsection{Correlation of four-degree-of-freedom modes}
\label{sec:4D_corr}

\begin{figure}[H]
    \centering
    \begin{adjustbox}{width=0.7\paperwidth, center}
    \begin{subfigure}[b]{0.49\textwidth}
        \includegraphics[width=.90\textwidth]{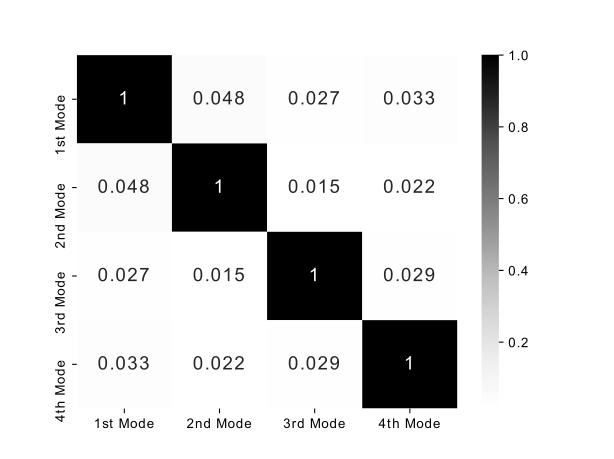}
    \end{subfigure}
    \begin{subfigure}[b]{0.49\textwidth}
        \includegraphics[width=.90\textwidth]{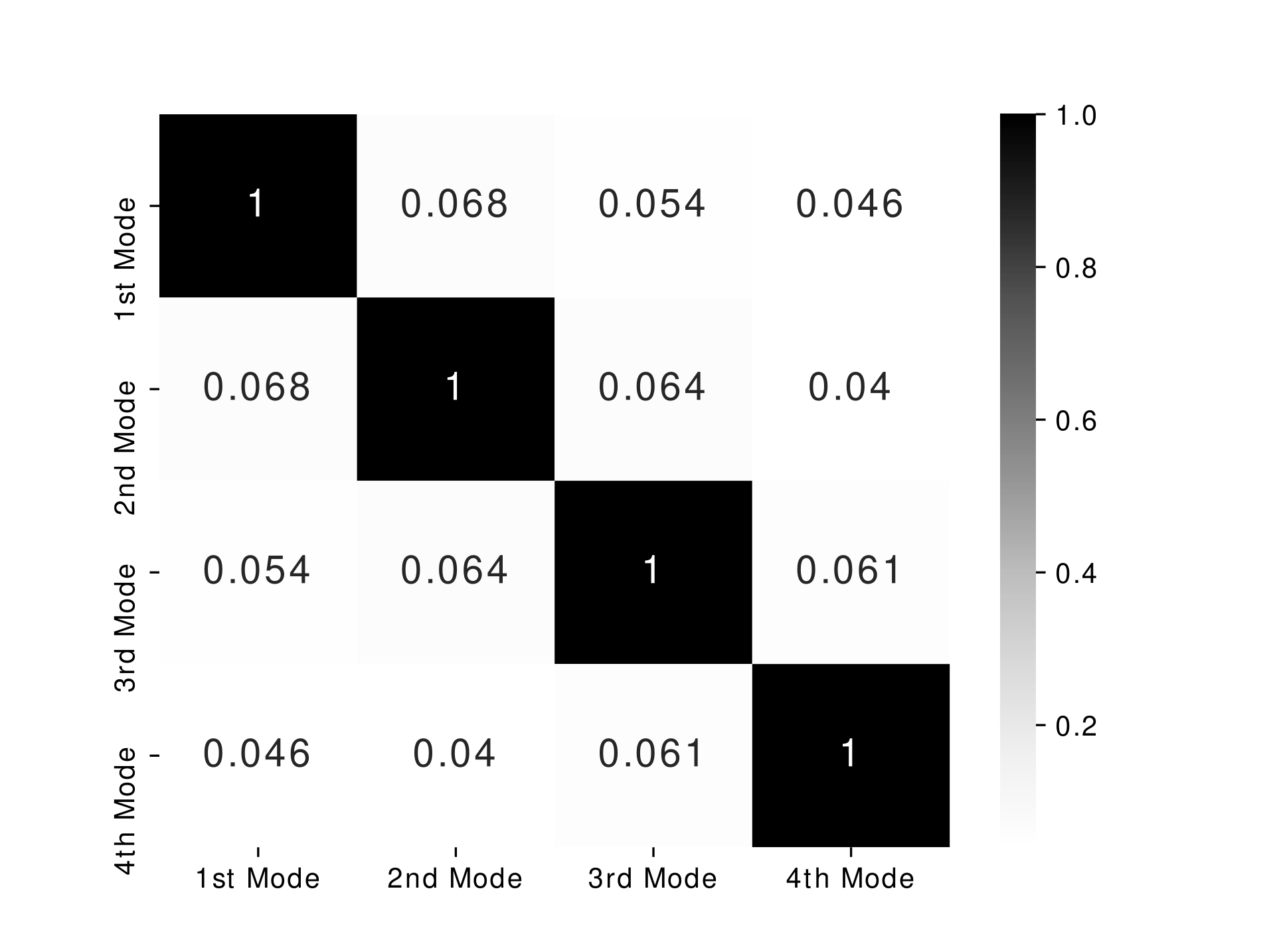}
    \end{subfigure}
    \end{adjustbox}
    \caption{Pearson's linear correlation coefficient (left) and distance correlation coefficient (right) of the modal decomposition computed for the four-degree-of-freedom system.}
    \label{fig:4d_corrs}
\end{figure}

For the four-degree-of-freedom system, the correlation coefficients are shown in Figure \ref{fig:4d_corrs}. The correlation coefficients for this case also appear to be low enough to assume that the modal coordinates are basically uncorrelated.

\subsubsection{Experimental data: State 12}

\begin{figure}[H]
    \centering
    \begin{adjustbox}{width=0.7\paperwidth, center}
    \begin{subfigure}[b]{0.49\textwidth}
        \includegraphics[width=.90\textwidth]{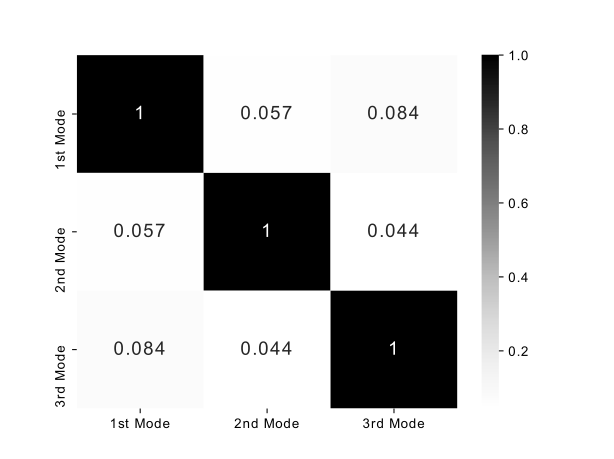}
    \end{subfigure}
    \begin{subfigure}[b]{0.49\textwidth}
        \includegraphics[width=.90\textwidth]{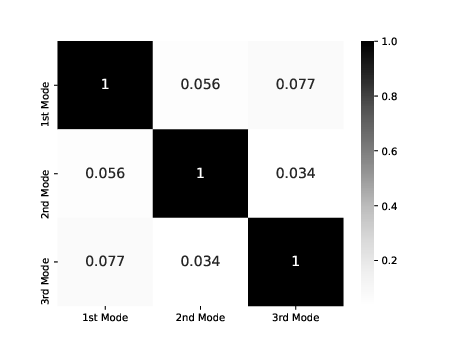}
    \end{subfigure}
    \end{adjustbox}
    \caption{Pearson's linear correlation coefficient (left) and distance correlation coefficient (right) of the modal decomposition computed for the experimental system.}
    \label{fig:lanl_corrs}
\end{figure}

For the State 12 of the experimental system, the correlation coefficients are shown in Figure \ref{fig:lanl_corrs}. Again, the distance correlation and Pearson's correlation coefficient have low values.

\subsubsection{Experimental data: State 14}

\begin{figure}[H]
    \centering
    \begin{adjustbox}{width=0.7\paperwidth, center}
    \begin{subfigure}[b]{0.49\textwidth}
        \includegraphics[width=.90\textwidth]{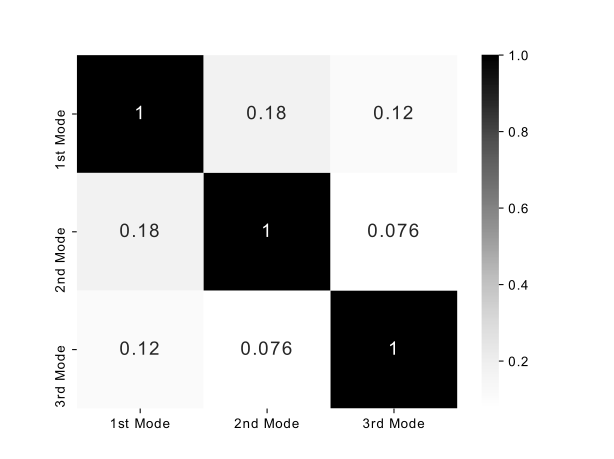}
    \end{subfigure}
    \begin{subfigure}[b]{0.49\textwidth}
        \includegraphics[width=.90\textwidth]{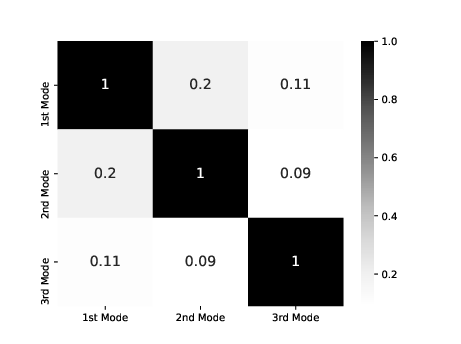}
    \end{subfigure}
    \end{adjustbox}
    \caption{Pearson's linear correlation coefficient (left) and distance correlation coefficient (right) of the modal decomposition computed for the experimental system.}
    \label{fig:lanl_corrs_state_14}
\end{figure}

For the State 14 of the experimental system, the correlation coefficients are shown in Figure \ref{fig:lanl_corrs_state_14}. This time the correlation values are higher than before; this might also be the result of the highly-nonlinear system. The highest correlation is observed between the first and the second modal coordinates. It can also be explained by energy observed in the PSD of the second modal coordinate (Figure \ref{fig:PCA_CG_LANL_PSD_state_14}) close to the frequencies of the first mode.

\section{Conclusions}
\label{sec:conclusions}

An algorithm was described as an alternative of the framework developed in \cite{worden2017machine}. The algorithm aims at performing data-driven nonlinear modal analysis. The concept on which the nonlinear modal analysis is based is that of decomposition into SDOF systems. However, instead of decomposing the target dataset of displacements (or accelerations) into a modal space, a mapping from a pre-defined modal space onto the dataset is attempted. Both forward and backward mappings are learnt using a cycle-GAN model. In contrast to the classic application of such a model, a restriction of orthogonality is also imposed here. This restriction corresponds to the orthogonality of the mode shapes. Moreover, a model selection criterion different to the loss function used during training, based on the inner product of two PSDs, is used to pick the model that separates the modes most efficiently. 

The selection criterion, apart from selecting the model that best separates the modes, is also expected to implicitly minimise the correlation of the modes. Given that a modal variable reacts only to some interval of frequencies that is not overlapping with the rest of the modal coordinates, its statistical correlation with the rest of the modal variables is minimised. The minimisation of the correlation is also implicitly imposed via the adversarial training. The target distribution of the generator, which transforms from natural coordinates to modal, is a multivariate Gaussian with diagonal correlation matrix. The adversarial training is forcing the generator to generate samples with similar distribution and therefore with correlation between modal coordinates close to zero.

Via three simulated case studies and two experimental, the performance of the algorithm was illustrated. The proposed model was able to decouple modes in two, three and four degree-of-freedom systems with cubic stiffening nonlinearities and also in two bilinear (bumper column) experimental cases. The main reason that the algorithm performs better than the algorithm proposed in \cite{worden2017machine} is probably that there are no restrictions on the polynomial order of the composition (or decomposition) function used or on the objective function. 

Parallel to the training of the decomposition, the superposition function is also trained and as illustrated, yields accurate results (below $2\%$ normalised mean square error, except for the second experiment, implying very good accuracy models); thus providing a solution to invertibility problems in \cite{worden2017machine}. The second generator can be used to map from the modal space to the natural coordinate space. The only case that a higher error in the superposition function was observed was the most nonlinear experimental case. The nonlinearity in the latter case is imposed by the bumper and the column defining a bilinear nonlinearity. When the bumper is too close to the column and too close to the equilibrium point, the nonlinearity is also imposed by the striking of the two elements, which causes sudden energy dissipation and non-proportional damping, as well as additional vibration components to the structure. Nonlinearity on non-proportional damping might require considering also the velocity variables within the algorithm, in order to achieve a more efficient decomposition. The problem might also be that the algorithm did not have enough data to satisfy all the constraints imposed during training (adversarial loss, orthogonality loss and reconstruction loss) given how highly and non-smoothly (sudden stiffness change and energy dissipation) nonlinear the case is. In any case, if the accuracy is not sufficient, one can train a separate regression algorithm (for example a Gaussian process or a regression neural network), in order to define a superposition function. 

Two correlation coefficients (Pearson's correlation coefficient and the distance correlation) are calculated for the modal coordinates that were computed by the algorithm. The values of the coefficients seem to confirm the initial assumption, that the \textit{a priori} selection of a modal space and the application of PCA on the physical coordinates forces the modal coordinates to be uncorrelated. Regarding the correlations, again in the most nonlinear case higher values of the correlation metrics were observed. The high correlation might be the result of lack of data as well as the largely nonlinear system in combination with a non-smooth nonlinearity. The distance correlation is used because Pearson's correlation can be `fooled' by nonlinear correlations. The distance correlation cannot be tricked by higher-order correlation and this is verified by the values on the off-diagonals that are generally higher than the corresponding Pearson's correlation.

In total, the proposed algorithm solves most of the problems that remained in \cite{worden2017machine}. The algorithm is able to perform for higher-degree-of-freedom systems than the one presented in \cite{worden2017machine} and also for the experimental data. Moreover, the dot product criterion presented replaces the `by eye' selection of the model that best separates the modes. The problem of superposition is also solved, since by using the cycle-GAN algorithm, the superposition function is trained in parallel. Finally, the modal coordinates seem to have low correlation values except for the harshly-nonlinear case, an issue that remains to be solved.

\section*{Acknowledgements}
This project has received funding from the European Union’s Horizon 2020 research and innovation programme under the Marie Skłodowska-Curie grant agreement No 764547. KW would like to thank the UK Engineering and Physical Sciences Research Council (EPSRC) for an Established Career Fellowship (EP/R003645/1). DW would like to acknowledge the support of EPSRC grand EP/R006768/1.

\bibliographystyle{elsarticle-num}
\bibliography{cycleGAN_nonlinear_modal_analysis}

\end{document}